# What do professional software developers need to know to succeed in an age of Artificial Intelligence?

Matthew Kam
Google
Mountain View, CA, USA
mattkam@google.com

Cody Miller
Google
San Francisco, CA, USA
codymi@google.com

Miaoxin Wang
Trilyon
Seattle, WA, USA
mxworking@gmail.com

Abey Tidwell
Google
San Francisco, CA, USA
atidwell@google.com

Irene A. Lee
Education Development Center
Waltham, MA, USA
ilee@edc.org

Joyce Malyn-Smith
Education Development Center
Waltham, MA, USA
jmsmith@edc.org

Beatriz Perret
Boston College
Chestnut Hill, MA, USA
beatriz.perretg@gmail.com

Vikram Tiwari
Assembled
San Francisco, CA, USA
hi@vikramtiwari.com

Joshua Kenitzer
Google
Buffalo, NY, USA
jkenitzer@google.com

Andrew Macvean
Google
Kirkland, WA, USA
amacvean@google.com

Erin Barrar
Google
Cambridge, MA, USA
erinbarrar@google.com

## Abstract

Generative AI is showing early evidence of productivity gains for software developers, but concerns persist regarding workforce disruption and deskilling. We describe our research with 21 developers at the cutting edge of using AI, summarizing 12 of their work goals we uncovered, together with 75 associated tasks and the skills & knowledge for each, illustrating how developers use AI at work. From all of these, we distilled our findings in the form of 5 insights. We found that the skills & knowledge to be a successful AI-enhanced developer are organized into four domains (using Generative AI effectively, core software engineering, adjacent engineering, and adjacent non-engineering) deployed at critical junctures throughout a 6-step task workflow. In order to "future proof" developers for this age of AI, on-the-job learning initiatives and computer science degree programs will need to target both "soft" skills and the technical skills & knowledge in all four domains to reskill, upskill and safeguard against deskilling.

## 1 Introduction

Generative Artificial Intelligence (GenAI) is disrupting both the profession and practice of software engineering. The popular press [8, 24, 26, 35] is replete with questions about the career relevance of learning to code when Large Language Models (LLMs) are demonstrating capabilities such as code generation. As of 2024, AI-powered developer tools such as GitHub Copilot are widely used by over 1.3 million paid users [34]. Randomized controlled trials of AI-powered developer tools in both lab [45] and real-world, industry work settings [9] show AI's promising impact on developer productivity. For instance, professional developers who use GitHub Copilot to carry out everyday, enterprise-level work for durations lasting between two and seven months exhibit a 26% productivity improvement as measured by the number of Pull Requests completed weekly [9]. In other words, for those developers whose responsibilities involve only coding, once augmented with AI, the same volume of work that used to take a business week can be performed in a 4-day workweek. More importantly, given AI's likelihood to generate code that needs to be reviewed and edited, increased coding speed is not attained at the expense of a detectable impact on code quality [9].

More troubling, however, the productivity gains from AI do not appear to be distributed equally among developers. A sparse body of research on the skills & knowledge to be a successful developer in the age of GenAI [2, 4] suggests that developers need prerequisite skills and knowledge in order to leverage AI tools productively. For instance, the most junior developers – with under one year of experience – take 7-10% longer to complete their tasks in some situations when using AI tools compared to not using AI [4]. The same study recommends that this segment of developers receive "additional coursework in foundational programming principles - for example, coding syntax, data structures, algorithms, design patterns, and debugging skills" [4] in order to use AI developer tools productively. Similarly, developers who are more skilled respectively at software engineering – including requirements engineering – are more successful at using Large Language Models (LLMs) for building production quality systems [43] and requirement elicitation tasks [2].

Specifically, our research questions are:

- **RQ1.** Among developers who are experts at using GenAI to accelerate their work, what are the tasks that they use GenAI for at work?
- **RQ2.** What skills & knowledge are essential for developers to effectively leverage GenAI for their tasks at work?

This paper contributes to our understanding of the skills & knowledge to be a successful developer when leveraging AI-powered developer tools. In the rest of this paper, we first survey industry sources and academic research on what is currently known about GenAI in software development. We then detail our research process for developing an "occupational profile" [51] of the AI-enhanced developer. This process (DACUM, short for "Developing A CurriculUM") has been widely used to create occupational profiles that are subsequently used for job analysis, workforce training, and the development of national occupational skill standards. Next, we report on 12 work-related goals that developers on the cutting edge of using AI-powered developer tools are using AI to accelerate in their jobs, together with 75 tasks that we discover they carry out with AI to achieve these 12 goals. We share 5 insights on how AI is reshaping the work of a developer, and the skills & knowledge needed to thrive in AI-enhanced software development. We show how these skills & knowledge are deployed in tandem along a 6-step workflow that applies to any of the 75 tasks. Finally, we discuss the implications for both higher education and on-the-job learning to future-proof developers for the age of GenAI.

As an international expert group across academia, industry and government notes: “A human-centered approach to AI that … interacts with individuals while *respecting human’s cognitive capacities* [our emphasis]” is one of the six grand challenges in human-centered AI [15]. In order to adopt a human-centered AI perspective on software engineering and software engineering education, we take a nuanced approach toward the developer user’s cognitive capacities by situating the skills & knowledge for the age of GenAI within the following three core principles:

- **Principle 1. Facilitate Reskilling:** Enable developers to adapt their skills & knowledge to new and changing tools, including AI-powered ones, in order to *maintain* their productivity.
- **Principle 2. Enable Upskilling:** Provide opportunities for developers to deepen their expertise and advance mastery of their craft, thereby *increasing* their productivity and promoting their career growth.
- **Principle 3. Safeguard Against Deskilling:** Prevent automation from displacing developers in the short-term at tasks associated with opportunities to develop the skills & knowledge essential for long-term growth and productivity [3]. This risk associated with "skill loss" and undermining junior workers from growing into a pipeline of expert workers applies to all industries impacted by AI, and is estimated to be a “multi-trillion-dollar problem” [3].

Our research contribution is significant in at least three ways. Firstly, with 63% of professional developers already using AI [49], the challenge in inventorying the use cases for AI tools – together with the skills & knowledge for success with these tools – necessitates a qualitative "deep dive" with a small sample of professional developers who are *experts* at AI-enhanced software development. Such experts are likely to be found among early adopters of AI tools. Below, we describe both our stringent criteria and process lasting five months for recruiting such expert informants. Our empirical research with a panel of experts adds to the validity of what individual expert accounts [33, 43] already report on the skills & knowledge crucial for AI-enhanced development. Secondly, existing curricula sources – such as those reviewed in Related Work below – are more likely to introduce the use of AI tools for writing code and tests, or selected stages in the software development lifecycle that are more suitable for teaching someone to become a software engineer. In contrast, this paper reports on 75 tasks – and their corresponding skills & knowledge – that span the *entire* lifecycle encompassing the conceptualization, development, deployment and maintenance of production systems. Thirdly, and along the same line, existing curricula sources mostly target pedagogical objectives without considerations about legacy code, for instance. In contrast, professional developers often work with existing, mature codebases. The skills & knowledge reported in this paper are comprehensive in accounting for what it takes to build software under a range of real-world, industry conditions.

## 2 Related Work

Recent studies have aimed to understand how and why professional developers use AI tools. A Stack Overflow survey found 63% of professional developers around the world currently use AI tools and, of those that use AI tools, the most common uses are to write code (82%), look for answers (68%), debug (57%), produce documentation (40%), generate assets & “synthetic data” (35%), understand code (31%) and test code (27%) [49]. Similarly, a survey with a convenient sample of 73 experts in Agile development methodologies uncovered a total of 17, 11 and 2 tasks respectively performed using AI that can be classified under hands-on technical work, engineering project management and team management [37]. Next, use cases that are more novel include using AI to arrive at system requirements [2], manage software development projects [7, 53], make context-aware changes to pasted code [17], migrate variable types in large codebases [38], and repair broken builds [23].

The reported benefits of using AI tools include higher task completion rates [12] and faster completion times for tasks – as much as 45-50% when documenting code and 20-30% when refactoring code [4] – and increased productivity as measured by weekly Pull Requests [9]. Importantly, increased coding speed was not attained at the expense of a detectable impact on code quality [4, 9]. Another reported benefit of AI tool use is that developers can shift their time to work on other, less repetitive tasks such as designing systems [28], meeting customer needs [4, 28], collaborating with team members [28] and learning [28]. Other benefits are seen in broader measures of developer productivity [16]: developers who use AI tools experience increased satisfaction [4, 12, 18], greater time spent on work that is more fulfilling [4, 12] and longer time in a state of "flow" [4, 12, 18] while expending less cognitive effort [12]. Beside productivity gains, accelerating the pace of learning is one of the most frequently cited benefits of AI tools, reported by 61% of professional developers [49].

Conversely, recent research also uncovered developers' insights on the limitations and potential harms of using AI for software development. Stack Overflow found that only a minority of professional developers concur that AI tools will increase code accuracy (29%), make workloads more manageable (24%), and improve collaboration (8%) [49]. They also found that fewer than 12% of developers fear job loss due to AI which aligns with other findings that 31% of developers question the accuracy of their existing AI tools, and nearly 45% of them believe their existing AI tools to be unsuitable for complex tasks [49]. Other frequently cited limitations of AI are its lack of understanding of context about the codebase, the organizational context, and inability to decompose a complex problem into smaller parts [4]. Furthermore, as a result of using GitHub Copilot, developers are spending less time on Stack Overflow, and have a decreased understanding about how and why the code works [5].

Given that AI's benefits and limitations depend on the capabilities of the actual AI tools, some research has been conducted to compare AI tools that are applicable to those stages of the software development lifecycle prior to system deployment (e.g. requirements, design, implementation) [27].

Currently, there is little research on the linkages between specific skills & knowledge and the effective use of GenAI in software development. McKinsey's research [4] finds that the most junior developers – those with less than one year of experience – take 7-10% more time to complete their work in some situations when they are using AI tools. The authors of the McKinsey study conclude that Generative AI tools are "only as good as the skills of the engineers using them" and "for developers to effectively use the technology to augment their daily work, they will likely need a combination of training and coaching" [4]. This finding is consistent with the prior observation that the developers who are more skilled at requirements gathering and validation are the same developers more likely to use LLMs effectively to elicit requirements [2]. Next, while these accounts do not constitute formal empirical research, there are nonetheless individual expert accounts of the skills & knowledge that are deemed crucial [25, 33, 43].

Prompt engineering features prominently in the skills & knowledge highlighted in existing work. For instance, prior work has contributed [39, 55] and evaluated [47] prompt patterns that developers can use to effectively leverage LLMs for tasks in the phases of the software development lifecycle prior to deployment (e.g. requirements elicitation, system design, implementation, refactoring). Where hands-on knowledge is concerned, recent Massive Open Online Courses and hands-on programming books on AI-enhanced software development introduce developers to LLM prompts for specific software development tasks [1, 14, 36, 39, 46, 50] across some phases in the development lifecycle [36, 46], selected GenAI tools for these tasks [1, 14, 39, 46, 50] and the fundamentals of machine learning (including GenAI) [13, 36, 50].

Yet, effective prompt engineering also entails understanding the attributes that make up quality code, and knowing how to prompt the AI tool for the right outputs to "actively iterate" on AI-generated code until it meets the desired quality [4, 10]. McKinsey's research concludes that developer training "should equip developers with an overview of generative AI risks, including any industry-specific data privacy or intellectual-property issues and best practices in reviewing AI-assisted code for design, functionality, complexity, coding standards, and quality, including how to discern good versus bad recommendations from the [AI] tools" [4]. That is, the effective use of AI in software development goes beyond knowing how to use LLMs.

## 3 Developing the Occupational Profile: Research Design

We chose to use the "Developing A CurriculUM" (DACUM) methodology [40, 42] that originated in the late 1960s [22] because it is a widely accepted method for job analysis, workforce training, and the development of national occupational skill standards. DACUM has been applied across more than 120 occupations in over 58 countries for more than 40 years [51]. The DACUM process rests upon three basic principles: 1) Expert workers can describe and define their jobs more accurately than anyone else; 2) An effective way to define a job is to precisely describe the tasks that expert workers perform; and 3) All tasks, in order to be performed correctly, demand certain knowledge, skills, resources, and behaviors. Thus a key first task is to identify and recruit a panel of 6-10 expert informants, which is the ideal number of informants required to run a DACUM workshop [52].

### 3.1 Expert Informants Recruitment

Our greatest challenge in defining a new or emerging occupation such as AI-enhanced software development was to identify and recruit those early adopter individuals where experts in the emerging field are more likely to be found. For this effort we sought expert informants who 1) were experienced developers on the cutting-edge of using AI tools to accelerate productivity at work; 2) came from workplaces where AI coding tools have been rolled out in teams across the organization, since collaboration is a significant contributor to complexity in software engineering [6]; 3) had extensive experience using GenAI coding tools day-to-day for at least 6 months; 4) submitted Pull Requests (PRs) several times per week

**11 Improve reliability to avoid production problems**

Google | EDC.ORG

| | Tasks | Examples | What the human does | What the AI does | What the human does | Cross-cutting SKAs | Specific Skills | Specific Knowledge | Specific Attributes | Tools |
|---|---|---|---|---|---|---|---|---|---|---|
| 11A | **Identify points of failure for critical components along system workflows and potential remediation items** | *Example: I give AI documentation or diagrams for the system or code itself and I ask AI to identify additional reliability considerations.*<br><br>*Example: I give AI a cloud architecture diagram and ask it to generate statements that instruct me to consider adding a failback database or a load balancer in another region to make sure there is some redundancy.* | Gives AI access to the **overall system architecture** and asks it to **identify critical components of the system that might fail** during high load or unforeseen events | AI assess these details to **identify** critical components and **might ask more questions to better understand the context** in which the architecture is running. It may give remediation items that can be used to optimize the infrastructure | Human **evaluates** these response and **iterates** with the model to get to a **better understanding** of the system components and either **apply** the suggested changes or **keep a note** of them in the failure guides. | | - Evaluate severity of the scalability and reliability weaknesses given the current state of the world (resource costs, DoS attack risk, product reputation, planned launches). | - Tech stack<br>- System architecture<br>- Design patterns<br>- Security flaws | - Observant<br>- Perfectionist<br>- Transparent | - LLM integrated resource planning tool (on cloud provider)<br>- LLM based mind mapping tool |
| 11B | **Add logging and monitoring to critical system flows** | *Example: I am working on code that is large and complex and I need to add log messages where relevant. I know what might go wrong. I ask the AI system to identify critical paths and apply a log message to that area, including creating the log message for me that will print out information when the code is executed.* | Gives AI access to the **code** and prompts it to **suggest what information to capture** and **where** to capture that information from. | AI gives a summary of what to capture and why. It also updates the code and adds log lines in the code | Human evaluates the LLMs output and determines if the suggestions are correct and if not, guides the LLM towards a better answer | | - Familiarity with the system and its expected behavior. | - Tech stack<br>- System architecture<br>- Design patterns | - Observant<br>- Transparent | - LLM Integrated code editor<br>- LLM integrated debugging tools |
| 11C | **Implement tracing (e.g., adding tracing calls to use Open Telemetry library) for critical flows** | *Example: I want to save guest info into the database in a hotel reservation system. Along the process, I don't know what might go wrong. I ask an LLM to generate a tracing component (for example, a piece of code from an API) that can check if hotel rooms are available while it is trying to save credit card information. Tracing is related to structured data formats.* | Gives AI access to the code and prompts it to update code to enable more detailed tracing for requests related to the CUJ. | AI updates code to enable the more detailed tracing and details how to access and interpret the results. | Human approves code updates, and takes code live either locally or in production. Tracing data is now available to both human and AI contributors to aid in future reliability optimization. | | - Familiarity with potential tracing options and how to interpret and act on the tracing data produced. | - Tech stack<br>- System architecture<br>- Design patterns | - Observant<br>- Time Consistent | - LLM Integrated code editor<br>- LLM Integrated tracing tool |
| 11D | **Plan for contingencies** | *Example: I need to plan for disaster, capacity, redundancy and rollbacks. I prompt AI to ask me what I need to consider my scenario. Say I am going to add a really big customer, AI helps look at logs and figure out what else I need to integrate the customer to my system. Ask LLMs what could possibly go wrong for a new client? What are the points of this plan and give me an outline to solve it.*<br><br>*Example: writing an emergency response playbook.* | Human prompts AI trained on internal resources for a risk analysis document or premortem related to a planned launch. | AI enumerates and ranks scenarios based on current code and documents as well as historical precedents and produces the requested document based on any identified plausible risks. | Human iterates on the document with further AI input, adjusting both launch plan and risk document. | | - Plan project<br>- Prioritize<br>- Determine trade-offs (prioritizing what to remediate after identifying points of failure)<br>- integrate human + AI knowledge | - Tech stack<br>- System architecture<br>- Tech debt | - Observant<br>- Perfectionist<br>- Systematic | - LLMs<br>- Collaboration tools<br>- Project mgmt tools<br>- Test environments |

**Figure 1: This Figure shows the portion of the occupational profile of the AI-enhanced software developer that corresponds to four tasks (e.g. identify points of failure for critical components) for achieving the goal "improve reliability to avoid production problems." More broadly, the occupational profile developed using the DACUM process uses a grid to show each of the 13 goals that developers use AI to perform at work, and the tasks that are performed to meet each goal. Each task is illustrated using example tasks as well as examples of what the developer does to interact with AI, what the AI does in response, and what the developer does in turn in response to the AI. These examples were gathered from both the panelists and advisors. For each of the tasks, the profile also presents a list of the specific skills, knowledge, attributes & tools necessary to succeed at the given task. Some of the skills, knowledge & attributes (SKAs) are cross-cutting and apply to all tasks. The cross-cutting SKAs are not detailed in this Figure and will be described more in Sections 4.2 and 4.3.**

that involved AI-generated code, and submitted PRs that have undergone significant prior iterations since the original AI-generated code; and 5) reported that AI has a "major" or "transformative" impact on their software development workflows, could share at least one use of AI coding tools that is novel (e.g. not reported in the literature), and contributed to their colleagues' understanding and use of AI tools through on-the-job mentoring or other technical leadership activities.

We set out to build a diverse panel of experts who represented various genders, industries, and company sizes. Informant recruitment painstakingly occurred over five months concurrently over three distinct channels and was non-blind. For our first channel, we enlisted C2 Research, an agency which specializes in recruiting research participants. C2 winnowed over 3,000 developers across industry down to 300 who expressed interest, had no conflicts of interest, and were available on the tentative workshop dates. We further winnowed down to 16 who appeared to meet the above five recruiting criteria, who were then phone screened by the 5th author, a researcher with a background in software development and 20 years of experience in Computer Science Education. 7 expert AI-enhanced developers were recruited from this channel and participated in the workshop. Our second channel, the Google Developer Experts (GDE) community [20] of more than 1,000 developers and technologists employed outside of Google who are thought leaders, influencers and experts on Google technologies yielded one additional US-based panelist. This panel of 8 experts from outside Google who were recruited through the first and second channels participated in the principal phase of the DACUM process (i.e. phase 2), which is the workshop where a first draft of the occupational profile of the AI-enhanced software developer was produced. (More details about all 3 DACUM phases are provided in the next subsection.)

Though extensive efforts were made for all channels to reach women who were developers experienced at using AI tools, we were unable to find enough of them who met all the five criteria above. In particular, of the three women who were phone screened, only one met all five criteria. Thus, while we sought greater gender diversity, our final panel comprised 7 males and 1 female. We believe this gender imbalance on the panel reflects a broader industry imbalance in gender diversity at the forefront of AI-use by developers. Recent research found that women comprise only 22% of AI talent globally, with representation at senior levels at less than 14% [44]. The gender disparity is extreme in the AI industry: women comprise only 15% of AI research staff at Facebook and 10% at Google [48]. The industries represented on the panel were ecommerce, cybersecurity, entertainment, food services, and scientific computing. In terms of company size, 2 panelists were from companies with between 1-5 employees; 1 panelist from a company with between 11-50 employees; 1 panelist from a company with between 200-500 employees; 1 panelist from a company with between 1000-3000 employees; and 2 panelists were from companies with over 10,000 employees.

Our third channel – in-house subject matter experts within Google – was used to identify and recruit 13 subject matter experts (all males) within Google who participated as advisors in both phase 1 (familiarization) and phase 3 (validation). We received 200 applications (176 males, 24 females), and winnowed down to 16 applicants who met most of the above screening criteria. In particular, we prioritized those who had at least 6 months of early adopter experience with AI coding tools outside of Google (in addition to those AI coding tools being designed and deployed within Google). Through our rigorous recruitment process with these three channels using the above criteria, we are confident that our 8 participating panelists and 13 advisors are *expert-level* AI-enhanced developers who could effectively describe a comprehensive set of work tasks in AI-enhanced software development and the skills, knowledge & dispositions needed to perform that work effectively.

## 3.2 Three-Phased Profile Development Process

DACUM involves three major phases: 1) exploring the target occupation; 2) producing an occupational profile [22, 41]; and 3) validating the profile.

Phase 1: Exploring prior to DACUM workshop. Phase 1's goal was to arm ourselves with as much understanding as possible — at a concrete level -– about AI-enhanced software development, before we set out to conduct the DACUM workshop (Phase 2). We drafted a straw definition of an AI-enhanced software developer drawn from prior research. We asked our 13 advisors to identify how they were already using AI to enhance business value in Google's 30 high-level software development goals. They provided rich details including screenshots showing exactly how they used AI to accelerate those goals, reflected on how AI has changed their work, and whether AI has been a net positive or negative change for their productivity.

Phase 2: DACUM workshop. Adapting the DACUM process which we used successfully in the past to develop profiles for emerging fields [11, 21, 29–32, 54], Phase 2 was a 2.5-day workshop held Friday - Sunday (May 31 - June 2, 2024), so that panelists do not have to take "paid time off" to participate in the research. The 8 panelists first reviewed and edited the straw occupational definition, then voted to identify 11 of the Google developer goals and 2 new goals they added to best describe their work as AI-enabled software developers. When vetted, these goals were found to be highly consistent with the same developer goals from Phase 1 that our advisors believed AI to be best positioned to enhance business value. Next, panelists broke down the 13 goals into smaller tasks and provided examples of what each task looks like when enabled with AI and then worked in pairs to list the knowledge, skills, attributes & tools necessary corresponding to each task. They presented their work for group consensus. The workshop ended with a focus group on how AI has impacted software development thus far and future trends. At the conclusion of the workshop, we had a draft occupational profile (figure 1) which included the occupational definition; a grid of 13 goals and their corresponding 91 tasks with examples; and for each of the 91 tasks, a list of the knowledge, skills, attributes, and tools necessary to succeed at the task. At the end of the workshop, panelists each agreed that the profile effectively captured their work as AI-enhanced developers.

Phase 3: Qualitatively validating the profile after DACUM workshop. Phase 3's goal was to ensure that the DACUM-styled profile, aimed at educators "Developing A CurriculUM", would also contain sufficient detail and be understood as the panelists intended by a technical audience that included professional software developers and computer science professors. The profile resulting from Phase 2 had some ambiguities because the 30 developer goals presented to panelists during the workshop contained ambiguous language which trickled down into the tasks, and because some language in the profile was meant to allow for flexibility in interpretation. As a result, we needed to support one uniform interpretation of all 13 goals by re-organizing some tasks and editing some of the technical language. For every statement in the profile that had multiple possible interpretations, we converged on the best interpretation with the advisors and 8th author (who was one of the original panelists and had context from earlier workshop discussions). Then we carefully and painstakingly edited the profile to include "signpost" language on the relevant "engineering parameter" (e.g. granularity, metrics, lifecycle stage, developer goal) where the panelists presumably intended flexibility of interpretation, so that the reader can more confidently exercise flexibility when interpreting the profile. In total, we proposed to reduce the total number of distinct tasks from 91 to 80 once there was consensus that 11 particular tasks contained sufficient duplication with other tasks. All panelists had the opportunity to "sign off" on the proposed edits before the profile was finalized on October 17, 2024.

# 4 Research Findings

## 4.1 Addressing RQ1: Among developers who are experts at using GenAI to accelerate their work, what are the tasks that they use GenAI for at work?

Among the 13 goals discovered during the DACUM workshop, in our professional opinion, the 13th goal (i.e. testing AI models) represents a highly specialized activity. Developers who need to *build* AI-powered software features – vis-a-vis *use* AI/LLM-based features to accelerate their work – are much more likely to undertake this goal. That is, success for this goal hinges on having the specialized skills & knowledge of an AI or Machine Learning (ML) engineer. Since this paper focuses on the broader practice of software development – whether or not the developer is an AI/ML specialist – we will not discuss this 13th goal in this paper (although this goal is found in the above occupational profile). A total of 75 tasks in the profile correspond to the 12 goals that we will discuss in the rest of this paper.

Specifically, the following 12 goals and pertinent[1] tasks in the profile apply widely to the practice of software development:

(1) **Contextualize a unit of work that needs to be done** comprehensively by leveraging AI to summarize voluminous user and stakeholder feedback, investigate competitors' offerings for strengths and gaps, and acquire more context about how the system behaviors they need to implement

[1]In the interest of space, we will not enumerate all of the 75 tasks that are performed to meet all 12 goals. Instead, we focus on describing those tasks that even an audience who's familiar with AI-enhanced software development are more likely to find novel.

fit into the overall system by prompting AI to explain the existing codebase and the locations in the overall codebase that relate to their work (e.g., file, lines of code) based on the context available to AI (8 AI-enhanced developer tasks in total for this goal)

(2) **Locate information related to this unit of work** (e.g. documentation, open source libraries, codelabs, API examples) by using AI to research the latest technologies for meeting the necessary technical requirements to deliver a compelling value proposition; locate and learn about the necessary technical knowledge, possible tools, and libraries for implementation (2 tasks)

(3) **Explore technical solutions for approaching this work** by leveraging AI to optimize for the given technical metrics (e.g. performance, complexity, cost, maintenance, reliability) and evaluate various technical options based on their advantages and downsides (8 tasks)

(4) **Develop and document a considered plan for completing this work** by using AI to help draft a high-level implementation roadmap, including the visuals, dependencies and other references (8 tasks)

(5) **Produce high-quality code** by using AI to remove complexity in the codebase and to convert design documentation – including visual schematics of how different components interact and how the user-interface ought to look – from idea to code (8 tasks)

(6) **Create and maintain holistic test coverage** by asking AI to evaluate test coverage, identify points of failure, recommend test cases, and generate test data that reflect real-world or stress-testing conditions (7 tasks)

(7) **Generate assets** (e.g., captions, images) by leveraging AI tools to research terminologies (e.g., styles, references) and optimize prompts iteratively based on various input and output modalities (e.g., reference image, text) (3 tasks)

(8) **Produce up-to-date documentation** by asking AI to generate accompanying inline code comments; assess necessary documentation changes; and generate the updated documentation, including user-facing examples (e.g., API tutorial for various use cases that a developer may need to implement) (4 tasks)

(9) **Ensure launch complies with legal, privacy and security** by leveraging AI to assess potential privacy violations, patent and/or copyright infringements, generate penetration tests, simulate those tests in order to identify anomalies, and propose possible solutions (7 tasks)

(10) **Monitor data and generate insights** for important events (e.g., changes in response time, error metrics post deployment, production interaction logs) by using a combination of GenAI and more traditional approaches (e.g. rule-based thresholds, statistical outlier detection, time-series analysis) to identify anomalies and generate predictions on the outcome of taking a given approach (9 tasks)

(11) **Investigate issues in production** by providing AI with code, system and error logs, stack traces and even user-interface outputs to conceptually determine the sources of the issues and isolate the responsible code and/or configurations (7 tasks)

(12) **Improve reliability to avoid production issues** by leveraging AI to analyze the codebase, proactively identify likely failure conditions, and recommend appropriate logging strategies, short-term mitigation measures, and long-term risk mitigation plans (4 tasks).

The above 12 goals and corresponding 75 tasks are carried out using AI tools that include LLMs embedded as features in popular developer tools such as GitHub Copilot, conversational LLMs like ChatGPT, multimodal LLMs, and other developer tools based on pre-GPT Machine Learning models.

Thinking about where the above 12 goals and 75 tasks are situated along the software development lifecycle provides us with an additional perspective for understanding how GenAI is changing the practice of software development. Specifically, in the lifecycle, Goals 1 to 4 primarily revolve around learning about the problem and design space, investigating technical options, and making decisions about these options prior to implementation. Goals 5 to 8 center around the production of high quality code, documentation, and other AI-generated artifacts. Goals 9 to 12 are about ensuring a compliant, robust production, launch and post-launch pipeline. To make all of the above clearer, we briefly summarize the key tasks that an expert-level AI-enhanced developer uses AI to carry out in each stage of the lifecycle, as further illustrated in Figure 2:

- **Plan (58 AI-enhanced developer tasks in total for this stage)**: engaging in the process of project-level [2], technical-level planning[3], and investigation[4] prior to detailed implementation.
- **Code (42 tasks)**: engaging in the process of generating code, developing tests, and configuring the necessary system setup.
- **Build (11 tasks)**: engaging in the process of transforming source code into a deployable artifact.
- **Test (28 tasks)**: engaging in the process of evaluating the system at various levels and using multiple types of testing to ensure quality.
- **Release (8 tasks)**: engaging in the process of making a version of the system available.
- **Deploy (7 tasks)**: engaging in the process of making a system operational in a specific environment.
- **Operate (27 tasks)**: engaging in the ongoing process of keeping a system running smoothly. It overlaps with other stages when tasks involve coordination and task management work. One example task is triaging active production issues which is also mapped to both the "Operate" and "Plan" stages.

[2]Project planning in this paper refers to the overall management and execution planning of the project. The process of defining the scope, objectives, and resources required, estimating effort and resources, and risk management.

[3]By technical planning, we mean the process of defining the technical blueprint of the system. To have a detailed design plan, it requires parallel work in "solution investigation" together to explore implementation options, comparing them, and selecting the optimal solution based on various constraints and trade-offs. Comparing to "solution investigation", this stage focuses on the "what" and "why"

[4]The solution investigation process bridges the gap between the "what" (the high-level tech and project plan) and the "how" (the actual code). It's where the technical details are fleshed out and the feasibility of different approaches is thoroughly examined. This stage emphasizes practical experimentation, prototyping, and risk assessment.

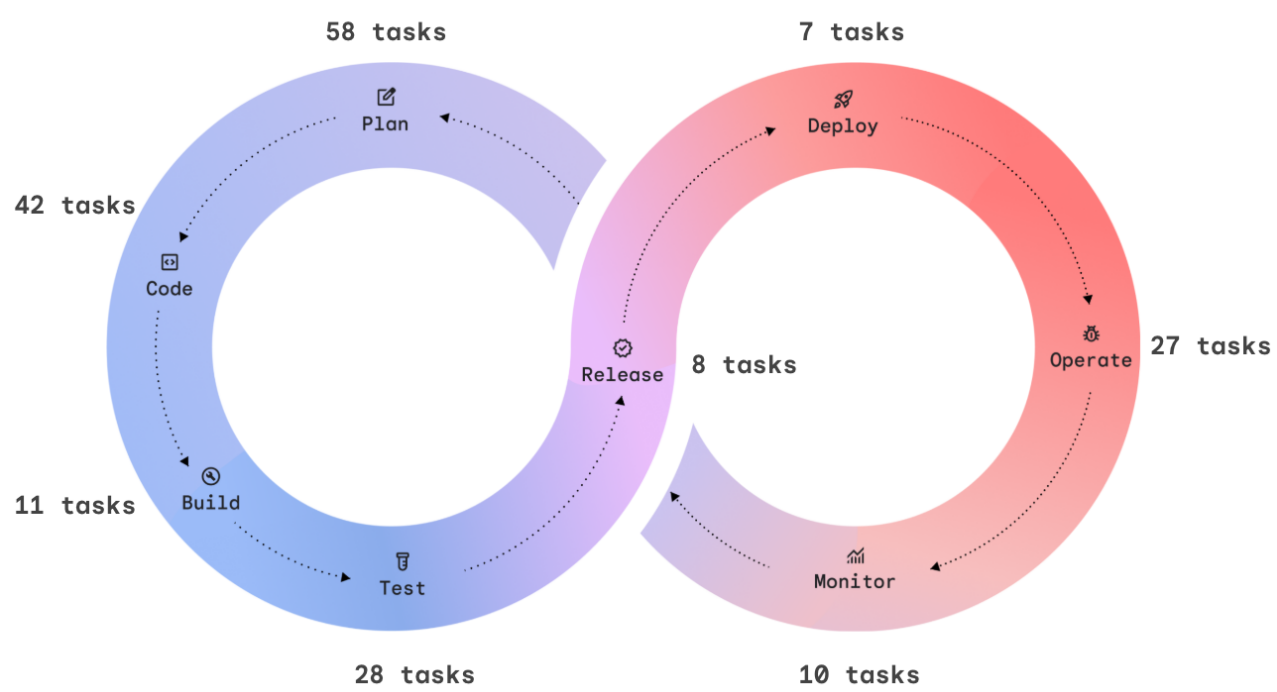

**Figure 2: One way to better understand and appreciate how AI is changing the practice of software development is to consider which of the 75 AI-enabled developer tasks apply to each stage of the software development lifecycle. The number of tasks across all stages total to more than 75 because the tasks are not mutually exclusive. That is, several tasks are applicable to more than one stage. One insight that this lifecycle-based visualization offers is that as much as the literature reports on how developers are benefiting from GenAI for coding and testing, developers on the cutting edge of using AI are carrying out even more tasks in the "plan" stage with the help of AI.**

- **Monitor (10 tasks)**: engaging in the ongoing process of observing and tracking metrics (e.g., availability, error, performance, user behaviors).

With less time spent on repetitive work, AI frees developers to experiment – under shorter cycles – and come up with optimal solutions to problems. From the profile, we observe that developers seek to use AI to optimize for both business metrics (e.g. "selects one system that is most vetted, most efficient, cost effective and has better chance of success and profitability") and engineering metrics (e.g. "complexity, performance, maintenance") **(insight #1)**. In other words, the future of software development augmented using AI is likely to entail an increased emphasis on meeting the intersection of business and technical goals.

> **One workshop panelist**: *To me it's ... like we have a specific problem and we think we can get there and then [we think] ... of something else and - oh, I would love to do that. And now we have the time [to explore an alternate solution] because AI has reduced the amount of time spent on what we normally would do.*
> **Moderator**: *So it seems like you're saying rather than just find a solution, find the best solution.*
> **Panelist**: *Find the best because [yes] it's like [we find] a solution, but maybe there's a better one.*

In some situations, AI tools facilitate individual developers in meeting their team's greater good. For example, generating example code for tutorial documentation (task) when producing up-to-date documentation (goal) can be time-consuming in the short term. With AI, developers complete this task more efficiently, and potentially reduce disruptions and save even more time in the long term – for themselves and coworkers – when coworkers can find answers from better documentation instead of having to ask **(insight #2)**.

Finally, while early research indicates that AI's time savings promote collaboration [28], our research suggests that the reality is more complex **(insight #3)**. When AI helps with finding previous code authors and explaining code, there is a reduced need for direct communication. Panelists report spending less time communicating with others. Similarly, an advisor notes that unless AI is used with care (e.g. AI-generated summaries of emails), human interactions can be reduced to information exchanges devoid of the human touch. On the other hand, when AI saves time asking and answering simpler questions, AI creates the potential for human interactions to focus on more meaningful questions. The same advisor adds that AI can help to put people in touch: *"AI more like a 'trusted advisor' rather than an 'executive' would probably be more of a net positive: less 'Here is a summary of your coworkers doc' and more 'Hey @ldap, we think this doc written by @foo relates to your existing work and plans.'"*

### 4.2 Addressing RQ2: What skills & knowledge are essential for developers to effectively leverage GenAI for their tasks at work?

In this subsection, we present a bigger picture of what these skills & knowledge look like when aggregated across all 12 goals and 75 tasks. The full dataset of skills & knowledge from the occupational profile by goal or task can be found at arXiv.org. In summary, the skills & knowledge to be an effective developer in the age of GenAI can be organized under four distinct domains, as illustrated using a T-shape diagram in Figure 3:

**Domain 1. Generative AI Usage**: This domain revolves around effective engagement with AI tools, including being able to assess AI's outputs and calibrating interactions with LLMs to achieve desired outcomes (see section 4.3 for more details). This domain also includes foundational knowledge of machine learning; as well as AI tools' capability and constraints (e.g., AI failures, including how and why AI hallucinates), which enable developers to select appropriate tools for given tasks and avoid wasting time using AI due to unrealistic expectations about AI.

**Domain 2. Core Software Engineering**: This encompasses the essential skills and knowledge for software development, which remain crucial for making sound technical decisions in the age of GenAI. Three key aspects about this domain include 1) coding and testing proficiency – knowing what constitutes high-quality maintainable code, best practices in defensive coding, optimization, systematic testing approaches, and debugging skills with comprehension and critical thinking; 2) risk assessment for production readiness (e.g., being able to recognize potential technical compromises, vulnerabilities, and planning for contingencies); and 3) good system design (e.g., possessing a deep understanding of existing and alternative system architectures, design & system constraints, and carefully weighing the potential benefits and drawbacks of multiple design solutions to meet requirements).

**Domain 3. Adjacent Software Engineering**: Specialized subdomains within and closely related to software engineering e.g. cybersecurity, industry-specific regulations, and emerging technological trends. The exact topics depend on the developer's context.

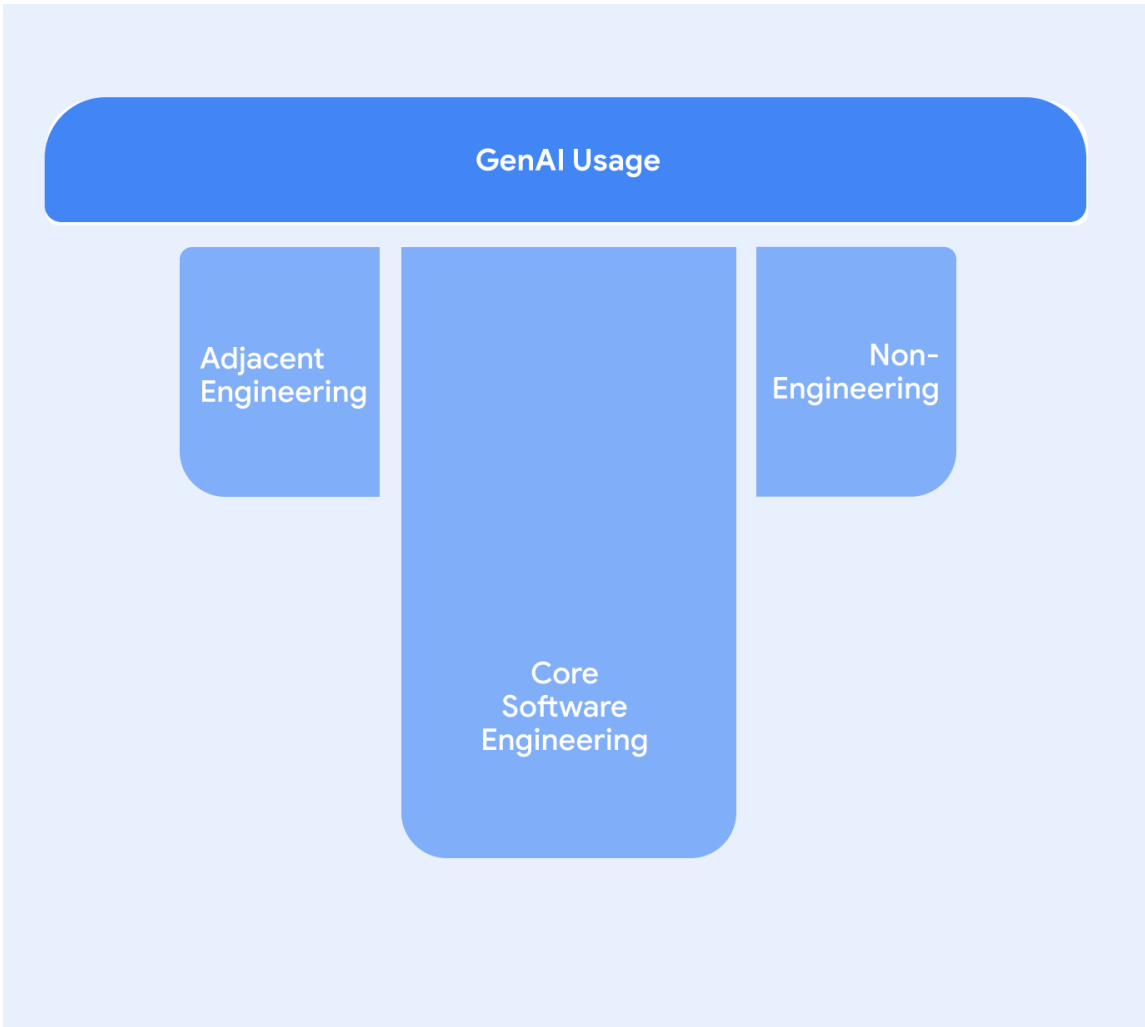

**Figure 3: The T-shape is a common visualization used to represent the skills & knowledge to be an expert worker in a field. The vertical bar of the "T" represents the deep expertise in a particular area that's core to success in the professional role. The horizontal bar in the "T" represents a wider breadth of knowledge & skills that complement the core, deep area of expertise. Knowing how to use GenAI effectively (dark blue) is the first knowledge & skills domain for the AI-enhanced developer to use AI productively at work, which in turn hinges on three additional domains (light blue). Specifically, the expertise of the AI-enhanced developer is characterized by depth in "core software engineering". This deep expertise is applied most effectively at work when complemented by breadth in adjacent domains that are both related and non-related to engineering. In practice, the exact skills & knowledge that goes into each of the four domains, and the relative importance – and size – of each domain, depends on organizational factors including how job roles are defined by a particular employer.**

For instance, in organizational settings that require software engineers to develop areas of specialization, a front-end developer focused on experimentation with proof-of-concept prototypes may not require knowledge of scale parameters for assessing and optimizing system resources for deployment planning.

**Domain 4. Adjacent Non-Software Engineering**: This encompasses at least 5 distinct sub-domains as we learn from the profile: end-user, customer, business or industry, competitors landscape, and market trends. Similar to Domain 3, the specific topics (e.g. General Data Protection Regulation) depend on the organizational context. For example, developers in smaller companies may need the skills & knowledge to "translate product (and technical) requirements into business needs (and vice versa)", "analyze feedback from users and internal stakeholders" and "integrate needs and feedback into work item". Developers in larger companies may focus more on contextualizing their work items so they are better able to evaluate trade-offs (e.g., cost-benefit analysis) and risks when optimizing for business value.

With the rise of AI, developers are changing how they acquire content knowledge **(insight #4)**. As one advisor puts it: *"Before GenAI... I'd have to maybe **read books** [respondent's emphasis]... While I [still] need to be familiar with other [programming] languages, [this] can be an issue for junior developers because I acquired this [knowledge]... by ... reading."*

## 4.3 A new workflow common across all tasks that weaves together the skills & knowledge

Regarding the skills & knowledge that are cross-cutting across all 75 tasks, we first observe that every task in the profile is carried out along a 6-step task workflow when the developer is collaborating with AI. In describing each of the 6 steps below, we use a color code to underscore where the corresponding skills & knowledge domains [5] essential for successfully completing the step are deployed in the workflow:

(1) **Identify**: Developers start an interaction by seeking and identifying the information that AI tools need (**Domain 1**), based on their knowledge of AI tools' capability (**Domain 1**) and the context of the task – from engineering (**Domain 2**, **Domain 3**) to business (**Domain 4**).
(2) **Engage**: They clearly express (**1**) a need (e.g., write a structured request that yields the specific information or product you are seeking) with the corresponding context (**2**, **3**, **4**), problem statement (**2**, **3**, **4**), and desired outcomes (**2**, **3**, **4**).
(3) **Evaluate**: They critically cross-check AI-generated artifacts, that may include content, designs, code, templates, or diagrams, for AI failures (**1**) against their domain knowledge (**2**, **3**, **4**). They verify for discrepancies against their expected end results, and other criteria for success (**2**, **3**, **4**).
When applicable, they put these artifacts in action (e.g., sandbox) to observe and compare. They note the limitations of the AI (**1**) and areas to improve its output (**2**, **3**, **4**).
(4) **Calibrate**: They steer AI tools toward the desired end results (**2**, **3**, **4**) with feedback (**1**) and additional context such as the the system (**2**), product (**4**), goals (**2**, **3**, **4**), and potential tradeoffs (**2**, **3**, **4**).
(5) **Tweak**: They take AI-generated artifacts and improve them based on their knowledge of the expected standards to meet (**2**, **3**, **4**).
(6) **Finalize**: Having arrived at the final version of the desired artifact, they produce up-to-date documentation (when applicable) with clarity and accuracy, such as reporting on how and why they arrived at the final version of the artifacts (**2**, **3**, **4**).

In practice, the steps in the above workflow may be *optional*, *iterative* and/or *non-linear*. Depending on the actual situation at work, when a developer carries out one of the 75 tasks with the help of AI, some of the above steps may be skipped (i.e. optional). Similarly, a developer may repeat a given step multiple times until the desired output is obtained from the AI (i.e. iterative). Finally, in order to be flexible based on what occurs during a given step, a developer may not perform the above workflow linearly from step 1 to step 6, and may need to "backtrack" to a previous step until the conditions are in place to advance to the next step (i.e. non-linear).

[5] The underlying skills & knowledge supporting interactions with AI are marked using the following color code: **Domain 1 Generative AI**, **Domain 2 Core software engineering**, **Domain 3 Adjacent engineering**, **Domain 4 Adjacent non-engineering**

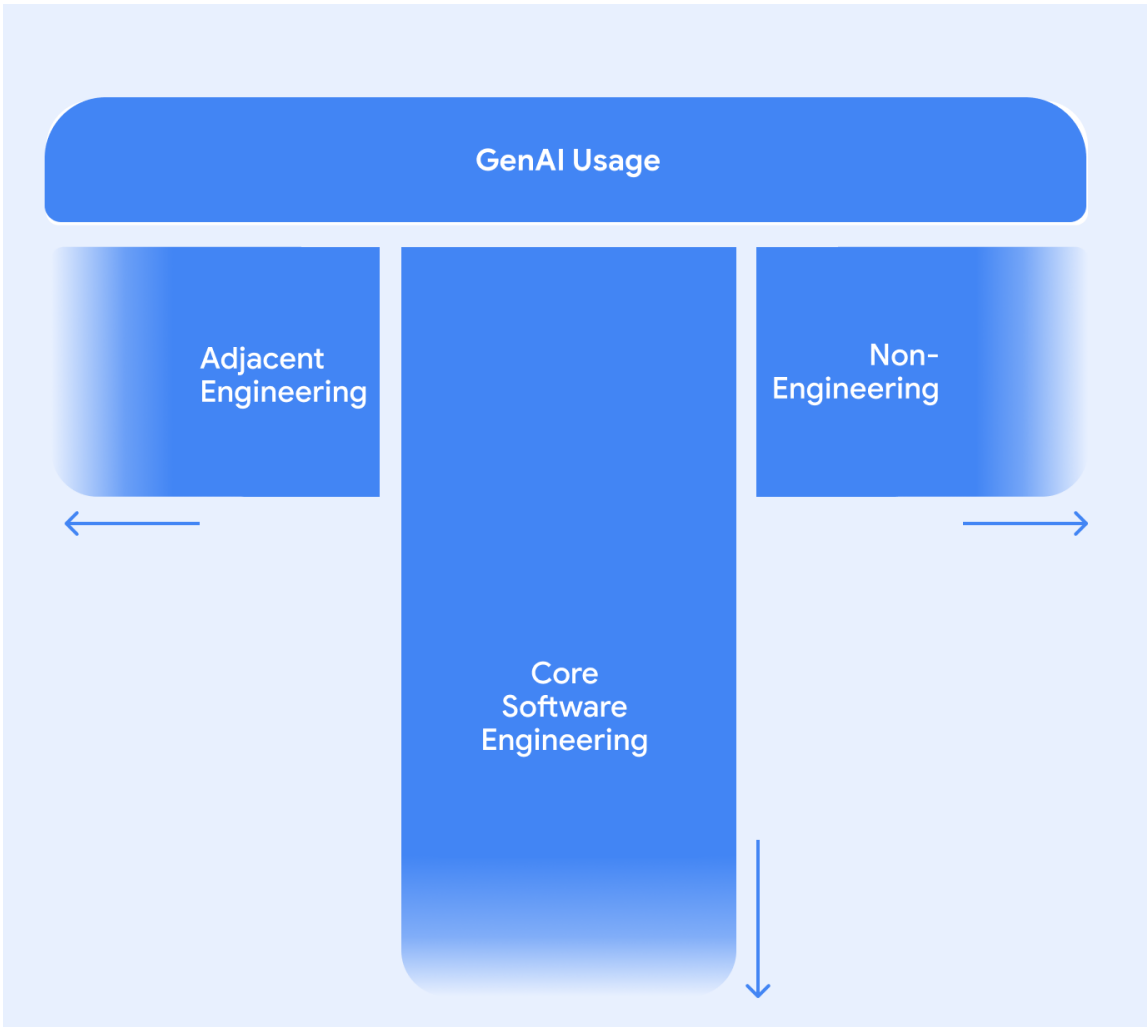

**Figure 4: The AI-enhanced developer (of tomorrow) will need to have the skills & knowledge of today's senior developer. From an upskilling perspective, developers will need to deepen their depth of skills & knowledge in the "core software engineering" domain, as well as widen their breadth of skills & knowledge in adjacent domains both related and non-related to engineering, in order to fully realize the benefits of AI tools while managing the risks.**

The use of GenAI tools is perhaps often understood to entail prompting an LLM only. Unfortunately, this corresponds to only a subset of the Engage step in the above 6-step task workflow.[6] The significance of the above task workflow is that it highlights the existence of other steps (e.g. Calibrate, Tweak) in an end-to-end workflow that are often essential for the successful completion of the overall, AI-enhanced task. Most importantly, the above description of the workflow shows where in the workflow, and how, Domains 2, 3 & 4 are absolutely critical for developers to effectively leverage AI tools. This perspective goes beyond prompt engineering in presenting a holistic understanding of the skills & knowledge that are critical for AI-enhanced software development. In fact, it calls for fluency in these skills & knowledge in order for the developer as a user of AI to be effective as the human in the loop, exercising appropriate control over AI throughout the workflow **(insight #5)**.

## 5 Implications

The media raises the specter of the human developer being displaced by AI. As reported in Related Work above, however, we are beginning to see the potentially transformative benefits of AI. As such, in our view, the perspective to take is less around the risk of AI displacing the developer, as much as the risk of the AI-enhanced developer displacing the developer who does not *reskill* to keep up with AI tools and be capable of realizing the maximum productivity benefits they offer. There is an utmost urgency for professional software developers to be equipped with the skills and knowledge to succeed in the age of GenAI, such that the human developer is capable of being in the loop at all times, ensuring that the benefits of AI are realized while its risks are managed. This starts with on-the-job learning opportunities as well as computer science degree programs going to the heart of where and how **GenAI usage (i.e. Domain 1)** appears throughout the above 6-step task workflow, reskilling developers for a changing toolchain and task workflow in which AI tools play an increasingly prominent role.

Next, when developers use AI to complete repetitive work with less time or effort, the time savings enable them to spend more time in the "plan" stage of the software development lifecycle (see Figure 2). As we have learned, AI-augmented tasks in the latter stage include understanding business and user requirements; translating these requirements into the corresponding technical requirements, metrics, and technical decisions; exploring alternative solutions (e.g. system design, architecture) for achieving better results on key metrics; and weighing possible solutions to arrive at the most suitable one. In this way, the AI-enhanced developer is delegating work that is more repetitive to AI, while spending more time in the role of a technical decision maker who reviews suggestions from AI.

Today, senior developers assume the entire range of the above responsibilities that include system design and making sound technical decisions, whereas junior developers are not ready to take on the fullest extent of these responsibilities until they *upskill* and acquire the depth of competency in **"core software engineering" (i.e. Domain 2)**. In other words, for a developer to fully realize the productivity benefits of AI, a regular AI-enhanced developer of tomorrow will need to have the skills & knowledge of today's senior developer (see Figure 4). There is an impetus for on-the-job learning opportunities as well as computer science degree programs to provide a deeper foundation in the skills & knowledge essential for good system design and technical decision making.

As we have also learned, developers receive support from AI to further their agency in going beyond solving engineering problems to designing and implementing the best solutions to business and user problems. In the process, developers cross domain boundaries. Using General Data Protection Regulation (GDPR) as an example of an adjacent, non-engineering domain; developers cross boundaries from core software engineering to an adjacent domain to build an implementation that is legally compliant. Specifically, they use AI to arrive at a basic understanding of the technical requirements, before they gain access to colleagues with expertise in the legal domain (e.g. in a big company setting) or in the absence of colleagues with legal expertise (e.g. in a small company setting).

The implication is that developers must *upskill* by also broadening their skills & knowledge in **adjacent domains related to engineering (i.e. Domain 3)** as well as **adjacent domains non-related to engineering (i.e. Domain 4)** (see again Figure 4). While computer science degree programs cannot feasibly teach everything, their breadth requirements can be reviewed to ensure that students have ample opportunities to acquire the necessary breadth of knowledge & skills both within, and more importantly, outside of engineering – aligned with Domain 4's five sub-domains above. This prepares students to meet the needs of the future that we cannot even imagine today.

[6] We would also note that as AI agents become increasingly popular in a developer's workflow, the human-AI interaction in the Engage step may evolve from how prompt engineering is currently done. Yet, our 6-step task workflow remains relevant in an agentic world.

In the age of GenAI, AI-enhanced developers who have stronger "soft skills" – in addition to the technical skills & knowledge we have just discussed – have a competitive edge over other developers. For example, strong communication skills enable developers to effectively seek help from coworkers once they have determined, based on their expertise with AI tools, that they have reached the limits of what AI can do to help with their tasks. Similarly, while AI can be double-edged sword in its impact on human relationships, on the whole, excellent collaboration skills enable developers to forge more meaningful relationships at work. Importantly, where instructional time is concerned, emphasizing soft skills does not need to come at the expense of technical skills & knowledge. Both on-the-job learning opportunities and computer science degree programs can investigate ways to better prioritize, structure and integrate supports for growing teamwork skills into existing curricula that target technical skills & knowledge.

Finally, AI fuels the impetus for the developer to be a continuous learner, especially of content knowledge. Users of LLMs need to continually evaluate the accuracy, relevance, etc. of the LLM's response. The discourse on AI literacy emphasizes the importance of skills such as critical thinking for recognizing hallucinations or misinformation. What is less often acknowledged or even recognized is that for users to critically evaluate the LLM's response, they need the necessary content knowledge [19] in AI, core software engineering, and adjacent domains for skills such as critical thinking to draw on and be successfully applied. Worse, with developers increasingly learning from AI-generated content instead of reading human sources (e.g. Stack Overflow [5]) whose content is more authoritative, the erosion of skills & knowledge appear to be beginning. To *safeguard against deskilling*, providers of both lifelong learning and higher education will need to consider more ways to increase access to high quality content.

## 6 Conclusions

Prior to the rise of AI, business-centered problem solving, collaborative teamwork, and iteration & experimentation are longstanding themes in software development. Similarly, developers already draw on skills & knowledge across domains whenever they can. What's new about AI is that technological disruption is amplifying these trends. Within the wider discourse on the automation of knowledge work, our research most strongly supports the view that AI is blurring the boundaries of what a professional developer does, in ways that augment rather than replace the developer. AI empowers the developer to design and build technology solutions that best solve problems for the user, customer and business. To the extent that the organizational context calls for the developer to become more business centric, work will likely transform to be even more collaborative, iterative, and experimentation-driven in pursuit of this aim.

Essential to the developer's success in this future of work is the ability to continually acquire new skills & knowledge —- including content knowledge. This knowledge base is multidimensional in that it spans how to effectively use LLMs at work, core engineering domains, as well as domains in and outside of engineering adjacent to core engineering. The multidimensional nature of this knowledge & skills supports developers in their work, where we see boundaries begin to blur in two ways. Firstly, AI is providing developers with the agency to more directly understand the business and user, and to more effectively incorporate these requirements into system designs at multiple levels of the technology stack. Secondly, developers where possible draw on this multidimensional knowledge base in a way that increasingly blurs the boundaries between their core engineering and adjacent domains.

## 7 Limitations

This study has four main limitations. The first limitation is the small sample size. The use of AI in software engineering was nascent at the time of the workshop, thus it was difficult to find expert informants who met our criteria for duration and depth of AI use. Of those we found, all were "early adopters" and of limited diversity and thus our findings may not be generalizable to all developers who eventually adopt AI tools for software engineering or as the field evolves. A second limitation is the inability to conduct a statistical validation of the findings. Attempts to do so were hampered by the limited number of potential validators. A third limitation is the dearth of prior research studies on this topic that would anchor our research in the past. Finally, a fourth limitation is that our study relies on self-report of AI use and was not independently verified. To mitigate potential sources of bias (such as selective memory, attribution bias, or exaggeration), we drilled down and asked informants to recount specific examples of AI tool use within each task. Despite these limitations, we argue that there is utility in this research; since software engineering is a field where rapid early adoption of AI is occurring, the findings from this study may offer important implications for on-the-job learning initiatives and degree programs related to software engineering; as well as new, potentially useful information for other fields that are moving forward with AI adoption.

## Acknowledgments

The first author expresses his gratitude to his family for their support when working on this paper. This research would not have been possible without the expert contributions of the 8 panelists and 13 advisors. We thank Allison Beck, Martina Blahova, Jacinda Ashley Jones, Shelly Bogetich Klose, Carmen Musat, Jeremy Neuner and CoreyMarie Roberts for their utmost assistance with expert informants recruitment; Megan Dizon, Aparna Kadakia and Laurie Wu for organizational support with the DACUM workshop; Joseph Ippito for help in planning and facilitating the workshop; James Lin and Mauli Pandey for their contributions to the literature search; and Monica Chan for her guidance with Overleaf. Finally, we thank Heather Breslow, Tao Dong, K.C. Geiger, Collin Green, Henning Fisher, Daye Nam, Yan Schober, Darla Sharp and Kevin Storer for constructive feedback on earlier drafts of this work; and Mohamed Dekhil, Maggie Johnson, Brian Saluzzo, Shivani Govil and Sara Ortloff for their executive sponsorship.

Occupational Profile

# AI-Enhanced Developer

The AI-Enhanced Developer (Software Engineer) works with, collaborates with, and orchestrates AI technology and systems to accelerate and more effectively gather information; plan and track work; develop, test, and commit high quality code; experiment with approaches; monitor the release of the code; and manage data.

Google | EDC.ORG

# Contents

# Introduction

# Adaptation of DACUM (Developing A CUrriculuM)

Step
**01**
Recruited AI-enhanced developers recognized as experts.

Step
**02**
Adhered to a valid and reliable methodology with 50+ years of evidence (DACUM).

Step
**03***
Developed an occupational definition for AI-enhanced software developers.

Step
**04**
Defined what AI-enhanced software developers do and need to know.

Step
**05***
Illustrated key tasks and required Skills, Knowledge and Attributes with examples.

Step
**06***
Described developer-AI interactions.

Step
**07**
Built consensus.

Step
**08**
Identified cross-cutting skills and knowledge.

Step
**09**
Produced a profile of the AI-enhanced software developer.

Step
**10**
Validated findings within Google.

*Adaptations made by EDC

# Key Contacts & Research Publication

**For details of the adapted DACUM process or our commentary on the AI-enhanced software developer described in this occupational profile:**

**Google Contact**
Matthew Kam
mattkam@google.com

**EDC Contact**
Joyce Malyn-Smith
jmsmith@edc.org

Occupational Profile

# Section 01

# AI-Enhanced Developer Occupation Definition

The AI-Enhanced Developer (Software Engineer) works with, collaborates with, and orchestrates AI technology and systems to accelerate and more effectively gather information; plan and track work; develop, test, and commit high quality code; experiment with approaches; monitor the release of the code; and manage data.

# Cross-cutting Skills

**Note:** Cross-cutting SKAs are essential for performing all tasks detailed in Section 02.

| Essential Skills | |
|---|---|
| **Identify** | Identify the information that AI-powered tools need, setting the stage for a successful developer-AI collaboration.<br><br>**1.** Given the context (e.g., technical, business) of the task and knowledge of AI tools' capability, identify the information / data sources to provide to the AI. |
| **Engage** | Initiate a collaboration with AI and clearly express the need (e.g., write a structured request that yields the specific information or product you are seeking).<br><br>**1.** Express ideas (e.g., in natural language, pseudo-code, code, and potentially with different modality and technical vocabulary)<br>**2.** Communicate clearly the context (e.g., technical, business), problem statement, and desired end-result |
| **Evaluate** | Critically assess AI-generated artifacts (e.g., content, design, code, template or diagrams) and put these artifacts in action (e.g., sandbox) to observe and compare.<br><br>**1.** Cross-check domain knowledge (e.g., software engineering, business) with AI-provided artifacts<br>**2.** Analyze data and patterns<br>**3.** Recognize AI hallucinations and other AI failures (e.g., incorrect solutions or AI's faulty assumptions)<br>**4.** Experiment with artifacts suggested by AI, including testing that it works correctly (i.e., put AI-generated artifacts in action (e.g., sandbox) to observe and compare)<br>**5.** Verify generated artifacts (based on the knowledge of expected result)<br>**6.** Discern limitations of the AI tool (by the errors it makes) based on the knowledge of association between errors to limitations (causes).<br>**7.** Identify areas for improvement in the solutions |
| **Calibrate** | Steer AI-powered tools to arrive at more useful info to rank AI-generated artifacts and make technical decisions.<br><br>**1.** Make a selection / best choice given human's domain knowledge (ranking implied) whether to adopt the AI-suggested artifacts or to continue improving it based on knowledge of system, product, goals, corp goals, tradeoffs*<br>**2.** Provide AI with feedback on its suggestions<br>**3.** Research the context, details, requirements and/or constraints for iterative refinement of the AI-suggested artif<br>**4.** Provide AI with additional details, requirements and/or constraints |
| **Tweak** | Take AI-generated artifacts and elevate them further.<br><br>**1.** Modify/adjust the AI-generated solutions based on their knowledge of standards and expectations |
| **Finalize** | Produce up-to-date documentation (when applicable) with clarity and accuracy once the final version of the desired artifact (e.g. design, code, content) is reached.<br><br>**1.** Produce up-to-date documentation |
| **Socialize** | Effectively obtain buy-in on their decisions on why their artifacts were created the way they were in socializing their artifacts, and **influence others in a way that ultimately** drives business impact.<br><br>**1.** Drive alignment within and between teams (on selection, choice, design etc.)<br>**2.** Manage stakeholders |

# Cross-cutting Knowledge and Attributes

**Note:** Cross-cutting SKAs are essential for performing all tasks detailed in Section 02.

| Knowledge |
| --- |
| Domain knowledge |
| Knowledge about errors, knowledge of how and why AI hallucinates |
| Knowledge of association between errors made by AI to AI's limitations (causes).<br>Example of errors: (1) not understanding that an LLM is viewing the full chat history whenever it makes a response, and that a response may be colored by earlier messages in the chat (2) missing awareness (indicators in the tool UI) that RAG or tool use has or has not been applied during the inference. |
| Pattern matching and data analytics |
| Knowledge of ground truth to verify accuracy |
| Knowledge of system, product goals, corporate goals, trade offs |
| Knowledge of expected outcome with proper assumptions |
| Knowledge of suitable prompt styles for different AI tools (Applicable to goal 10 only i.e. “Generate assets”) |

| Attributes | |
| --- | --- |
| Analytical | Patient |
| Articulate | Perseverant |
| Collaborative | Persuasive |
| Creative | Pragmatic |
| Curious | Proactive |
| Detail-oriented | Problem-Solver |
| Diligent | Skeptical |
| Discriminative | Technical |
| Inquisitive | Transparent |
| Leadership | Vigilant |
| Learning-oriented | User/Customer-first thinking |
| Logical | |

Occupational Profile

# Section 02

# 01. Produce high-quality code

**Note:** For each of the tasks in this section, the cross-cutting SKAs from the previous section apply, in addition to the SKAs specific to each task.

## 01. Produce high-quality code

**Note:** For each of the tasks in this section, the cross-cutting SKAs from the previous section apply, in addition to the SKAs specific to each task.

| | Tasks | Examples | What the human does | What the AI does | What the human does | Skills | Knowledge | Attributes | Tools |
|---|---|---|---|---|---|---|---|---|---|
| 1A | **Create a clear definition of the input-output formats** | *Example: I want to write a function to compute sentiment. I work with the model to figure out inputs and outputs. Outputs could be categorical, or numerical. Inputs might be just the plain email text, or can include the subject, or more.* | Human asks the model: "I want to build X, could you help me understand various inputs and outputs associated with this system?" | LLM details how these solutions are built, suggests existing solutions human can try, and suggests code samples. | Human iterates with AI to come to an understanding of what they would need to build X. | Check solution against broader problem scope. | - System architecture<br>- Limitations and boundaries on what the formats can be and can not be | - Observant | - LLM<br>- LMM |
| 1B | **Generate proof of concept** | *Example: I am trying to tackle a problem and would like a quick and simple implementation that solves the problem. I ask AI to assist in generating functioning code that illustrates how an end-to-end solution would work for a simple example.*<br><br>*Example: I am trying to determine how similar are two sentences. I ask the model to ""Write a function to compare two sentences using n-gram and return a similarity score.* | Given the **problem statement, input and desired output** I have a starter code. I ask AI to generate an end to end example for a particular use case that I'm looking for.<br><br>Context: The "end to end" focuses on the completeness of the solution. The reason "example" is used is to show that we might not use the exact code. It's a proof of concept. | AI generates an end to end example **for the particular use case** provided by the human. | Continues to ask AI to improve the proof of concept or generate other similar examples | | - Vocabulary<br>- Tech stack<br>- System architecture<br>- Limitations and boundaries on what the formats can be and can not be | | - LLM<br>- IDE with AI<br>- Sandbox environments |
| 1C | **Convert a loosely defined document (e.g., visual and technical designs) into code** | *Example:* ***I ask an LLM*** *to convert a design document for a system into the different components that a developer would need to build the system.*<br><br>*Example:* ***I ask AI*** *to convert a screenshot of a user interface into the code for that user interface.*<br><br>*Example: I ask AI to take in a diagram which describes adding a load balancer in front of VM instances on GCP. AI produces necessary commands.* | I ask AI to convert **a technical visual concept** or **a system design diagram** into code. | Depending on my technical/visual design input, AI converts it into **code** or a **breakdown of different components** needed for the project/problem statement. | Continue to iterate through AI outputs of the code until desired result | - Draw a rough sketch of the system | - Tech stack<br>- System architecture<br>- User experience | - Empathetic<br>- Quote: "Essentially the SWE has to build things for the end users, so she has to make sure that model's suggestions are relevant." | - Multimodal LLM<br>- Interface design tool integrated with LLM |
| 1D | **Convert thought process (logic) into code** | *Example: I want to solve a coding challenge and I have an approach I think I can use a graph to solve the question.* ***I ask AI*** *""Does it make sense to use a graph? and how"*<br><br>*Example: I have half of the code in my head and I write it, then* ***I ask AI*** *to complete it.*<br><br>*Example: If I ask AI to tell me the distance between earth and moon, AI cannot answer; but if* ***I ask AI*** *to write a function to compute it, AI can do this.*<br><br>*Context: this is "... a variation of [1C-1]. I have more defined requirements such as inputs and outputs"* | I provide a "seed concept" by **describing the inputs and outputs the desired code should have**, ideally with **specific names for the variables, functions, or object names** to steer the resulting code. | AI writes code that **follows the example patterns** provided by the human and **fills in the gaps with pseudo or functional code**. Usually by **extending** my idea from concept to proof of concept in a meaningful way. | Continue to iterate AI outputs **using higher quality examples of the desired output**, until the human decides the code is ready to test or use in production. | | - Vocabulary<br>- Tech stack | | - LLM<br>- IDE with AI |

# 01. Produce high-quality code

**Note:** For each of the tasks in this section, the cross-cutting SKAs from the previous section apply, in addition to the SKAs specific to each task.

| | Tasks | Examples | What the human does | What the AI does | What the human does | Skills | Knowledge | Attributes | Tools |
|---|---|---|---|---|---|---|---|---|---|
| 1E | **Generate pseudocode** | *Example: I am building a proof of concept system and want to focus on the critical coding parts.* ***I ask the AI*** *to generate the stubs of the code modules, allowing me to fill in the functionality. I* ***also ask the AI*** *to generate English pseudocode within the code modules to document and keep track of the system's operations.*<br><br>*Context: this is a variation of 1C-1. The logic is not well defined yet, and I'm in the process of breaking a real world problem into a computing problem. I'm trying to brainstorm the solution but in a code format (rather than one time solution). So in those cases, I would generally prefer to get a pseudo code and realize it's doable, rather than worry about syntax etc of a particular programming language. (I have a task in mind and want to see how can I solve a problem by writing a program.)* | I ask the AI to generate a outline or pseudocode **based on my system requirements**. | AI writes pseudocode with the basic functionality I described, or creates empty functions that appear to be appropriately named. | The human iterates with LLM to refine this until the logic 1) makes sense, 2) captures all the edge cases, and 3) still is simple enough to understand for the other developers (to ensure maintainability) | - Read pseudocode | - Software development<br>- System architecture<br>- Problem statement | | - LLM |
| 1F | **Minimize complexities by reusing and refactoring code** | *Example:* ***I ask AI*** *to scan the code and create a model with comprehensive knowledge of the codebase. I then query this model for existing functionalities to avoid reinventing the wheel.*<br><br>*Example: I provide AI with my code and ask it how to make the code simpler.* | The human **allows access to the code** or **provides the isolated changes** being reviewed in isolation along with the question, "Please review the following code and **suggest any simplifications, either by using new libraries, changing the logic flow or using different built-in functions**."<br><br>Variation: AI has access to the codebase. **As user starts writing the code**, the model generates (autocompletes) the code using the existing libraries already present in the code. | The AI provides **a list of changes** and **explains why** they are simpler or more efficient, in each case. | The human reviews the suggestions and **decides whether to implement** each change based on their **personal cost/benefit analysis of code readability/re-use/library bloat/execution speed.** | - Describe architecture and necessary functionality. *Context about why they need to be able to describe: As good as these models are, they don't have infinite context. Which means they either compress the information (in this case existing codebase) or the reduce the amount of data they focus on. So often, in codebases (since they are generally large), the model might miss a specific detail (for example, missing the detail about no-wifi connections) and produce code refactors that are not well suited for your task. If you know and can describe your codebase, you can provide that bit of missing/specific information to the model.*<br>- Identifying when code/functions need to be refactored. | - Feasibility of desired output<br>- Programming fundamentals<br>- When existing solutions may solve common tasks more accurately and tersely | **-** Imaginative *(being able to imagine what the end state of the system will look like, after the suggested refactor)*<br>- Perfectionist | - IDE to test AI generated code *[Context: during refactors (and reuses), existing tests must pass for the PR to be approved. The goal here is to evaluate the generated code, not to add new test cases (which is covered in Goal 5)]*<br>- LLM |
| 1G | **Produce extensible, readable code that is well-commented** | *Example: I need to deliver code that's reusable and understandable but oftentimes I have delivered only functional code due to time constraints.* ***I ask AI*** *tools within my IDE to access standardized style guides and can then refactor code to follow conventions for readability and clarity. These same tools also understand the entire codebase and can automatically generate comments and documentation.* | Human asks the model to write the code (optional: in the specified programming language) while providing all the contextual information (i.e. pseudo code, function definition and description etc). | LLM outputs the code that it has tested in a sandbox to make sure it works. | Human iterates over the code until it follow best practices. Generally LLMs take care of this since they are trained on curated datasets. | | - Tech stack | | - LLM<br>- IDE with AI |

**Note:** For each of the tasks in this section, the cross-cutting SKAs from the previous section apply, in addition to the SKAs specific to each task.

| | Tasks | Examples | What the human does | What the AI does | What the human does | Skills | Knowledge | Attributes | Tools |
|---|---|---|---|---|---|---|---|---|---|
| 1H | **Proofread code and suggests improvements with explanation** | *Example: I write a piece of code to convert from one data format to another.* ***I ask AI*** *to look at the code and suggest improvements. It responds with updated code that has better variable names, more efficient programming practices and up to date functions and syntax. It also explains what changes it made and why.* | When humans gets a PR to review, and doesn't understand what a tricky piece of code is doing, then can pass in that code and chat with the model to better understand what's going on. | AI explains the code and user asks more questions until they have satisfied their curiosity. | Human decides that the code should be simpler, asks the AI to generate a more readable version of the code and suggests it as a possible change in the PR review. | | - Tech stack<br>- Programming language<br>- Codebase | | - LLM<br>- IDE with AI |

# 02. Explore technical solutions (e.g., bugs, design)

**Note:** For each of the tasks in this section, the cross-cutting SKAs from the previous section apply, in addition to the SKAs specific to each task.

## 02. Explore technical solutions (e.g., bugs, design)

**Note:** For each of the tasks in this section, the cross-cutting SKAs from the previous section apply, in addition to the SKAs specific to each task.

| | Tasks | Examples | What the human does | What the AI does | What the human does | Skills | Knowledge | Attributes | Tools |
|---|---|---|---|---|---|---|---|---|---|
| 2A | **Identify problems (e.g., potential causes of errors) to solve** | *Example: When starting my research to solve a particular issue in the code, instead of searching the web to understand server messages or log messages, I give an LLM the context of my code and the error messages I'm getting and **ask it to** identify possible issues.*<br><br>*[context: working on existing code base]* | The human prompts AI with **specific log messages, error messages, or broken lines of code, giving as little or as much context as they feel should be necessary** to pinpoint or disambiguate potential answers. The human may request the AI responds with multiple potential causes/fixes with respective likelihoods/confidences | The AI responds with **one or many known issues that have been raised or documented that match the prompt** and suggest solutions to each issue. | The human now reviews suggestions and uses their broader context and understanding of the domain to **detect incorrect solutions or incorrect assumptions of the problem space**.<br><br>Potentially further **refine the suggestions by providing feedback on each one that was incorrect or providing further context on the one that seemed to be heading in the right direction** to clarify if it is still a good match (eg. provide library version information) | - Locating the correct logs in a terminal, IDE, or web dashboard | - Error log messages | - Observant | - LLM<br>- IDE |
| 2B | **Acquire necessary background technical knowledge for the problem and solution space** | *Example: When tackling a broad problem there's usually a need to know the correct terms and concepts. **I provide an LLM with metaphors and colloquialisms**. LLMs can help **guide me towards the needed vocabulary and concepts** to explore multiple solutions. [Context: this might be more relevant to developers who lack the prior knowledge necessary to formulate the problem and might be completely unaware of the solution space.]* | The human gives AI **context of application** or **domain** by **giving access to code or describing the tooling**.<br><br>Next, a prompt is given to request multiple possible **popular** solutions to solve a problem.<br><br>The prompt may include a **request** to provide a **glossary of terms** or **a link to documentation** for each implementation suggested. | The AI **provides a series of solutions** to the features/issue described in the prompt and returns relevant glossary of terms and/or documentation, as applicable | The human reviews the solutions, possibly **reading external documentation** and **learns more broad context as well as specific detail on the features and implementation details**.<br><br>The human may **double-check the suitability** of the solutions and use the broader language shared across each to **research the popularity/efficacy/ease of each** by searching **authoritative** sources (eg. npm/github) or **prompting further** to explore the same. | | - Required frameworks<br>- Potential solutions | | - LLM<br>- IDE |
| 2C | **Query internals and external solutions** | *Example: I ask AI to gather internal data from documentation and knowledge bases, and external solutions from sources like Stack Overflow. After building the model with this data, AI creates a knowledge graph to guide me to the final solution.* | The human asks the LLM how one can deploy a solution on the internal infrastructure. | AI uses the internal architecture documents and other relevant sources to come up with answers. It might also suggest generic solutions if it can't find anything internal | The human reviews the suggestions, then determines if that works for one's use cases | | - Tech stack | | - LLM<br>- IDE with AI<br>- task specific AI model |

## 02. Explore technical solutions (e.g., bugs, design)

**Note:** For each of the tasks in this section, the cross-cutting SKAs from the previous section apply, in addition to the SKAs specific to each task.

| | Tasks | Examples | What the human does | What the AI does | What the human does | Skills | Knowledge | Attributes | Tools |
|---|---|---|---|---|---|---|---|---|---|
| 2D | **Explore alternative approaches (including dependencies) to achieve better results on key metrics (e.g., complexity, performance, maintenance, cost)** | *Example: I need to write code and **I use an AI system** trained on internal frameworks to identify if there's already an internal solution.*<br><br>*Example: I need to perform a task and have options to solve it with services I already use. I ask an LLM whether there are better ways to write this exact code with or without the service to which I have access.*<br><br>*(Additional example and context for the task description: has it been done before? is this the only way I do it? 3rd party vs internal? questions existing services used: 3rd party vs. existing architecture; libraries)*<br><br>*[Context: variation from task 2A is that developers have the prior knowledge necessary to formulate the problem and are aware of at least one solution.]*<br><br>*Example: Alice wants to port their game from Android to PC, she has ported games before but wasn't satisfied with the tooling she used previously. She chooses to ask a general-purpose ChatBot to explore alternative technical solutions.*<br>***Alice**: I want to port an Android game to PC. Previously I used OpenGL, can you explain to me some alternative approaches and libraries I could use that might be better?*<br>***ChatBot**: Absolutely, let's explore some alternative approaches and libraries that might be a great fit for your Android game port to PC, considering your OpenGL experience. Alternative Rendering Approaches and Libraries: Vulkan [...]*<br>***Alice**: Vulkan sounds interesting, how do the concepts I know from OpenGL translate to Vulkan. What are the main differences I need to look out for?*<br>***ChatBot**: Let's map some core OpenGL concepts to Vulkan and highlight the key differences you'll encounter when making the transition. Conceptual Translation: OpenGL to Vulkan. OpenGL Concept, Vulkan Equivalent, Key Differences, etc..*<br>***Alice**: Can you write me a minimal Vulkan example in C++ so I can start playing around with it?*<br>***ChatBot**: Okay, here's a minimal Vulkan example in C++ to get you started. [...]*<br>***Alice** compiles the example and starts playing around with it in order to decide if switching from OpenGL to Vulkan is a worthwhile investment.* | The human **provides AI with a suggested solution** to a problem along with the **associated code changes** and information on the **libraries** and **services used**.<br><br>In addition to this, the question is posed, "Could I be doing this either in a **less complicated manner or in a more efficient manner**? you do not have to use the current libraries or architecture and are free to add any where necessary" | The AI provides solution(s) that **achieve the same goal but end up with fewer lines of code, fewer dependencies or a better solution by some metric (execution speed / future-proofing / cost)** | The human chooses **one or more of the suggestions** to use as an alternative and **learns more or tries to immediately implement to test efficacy.**<br><br>More prompting could be necessary if any answers don't give enough clarity or if answers are a bad fit, in which case a correction or additional context can be provided. | - Estimate time saved when solution is applied across codebase. | - Project, team, org, and corporate metrics and goals<br>- Programming Fundamentals | | - LLM<br>- IDE |

## 02. Explore technical solutions (e.g., bugs, design)

**Note:** For each of the tasks in this section, the cross-cutting SKAs from the previous section apply, in addition to the SKAs specific to each task.

| | Tasks | Examples | What the human does | What the AI does | What the human does | Skills | Knowledge | Attributes | Tools |
|---|---|---|---|---|---|---|---|---|---|
| 2E | **Suggest improvements to existing code and/or solution (e.g., improve efficiencies, additional features)** | *Example: I have perfectly working code and I can't think about what else to add to it, **so I prompt an LLM** with the existing code and **ask it** if there's something else I should add to it.*<br><br>*Example: I provide AI with my code and ask it how to make the code more efficient.* | The human asks LLM to look for efficiencies and edge cases in the code and suggest improvements. | AI looks at the code and suggests various possible efficiencies. Speed, maintainability etc | The human reviews the code and iterates with the model to ensure that all the requirements are met and the code/solution is not over complicated. | - Distinguish between features that actually add value for the target audience vs bloatware. | - Focus of product | - Perfectionist | - LLM<br>- IDE |
| 2F | **Determine where the solution lives within the "existing" application stack** | *Example: Use AI to isolate the relevant parts of the application stack where the eventual solution will live. Possible locations: front end, back end, architecture* | The human provides a context of the **platform / architecture** and describes the **desired outcome**. They then pose the question of **where best to situate** the solution and the **pros/cons** of each, depending on the desired optimization (eg. user speed/code complexity) | The AI will suggest implementations to achieve the desired outcome with additions or changes to existing architecture and the merits of each approach. | The human gains a better understanding of potentially new ways to solve problems, new uses for existing architecture and new architecture. This means a better solution is potentially proposed and/or the human's knowledge and aptitude around tooling is improved. | | - Tech stack<br>- Software product design<br>- System architecture | | - LLM |
| 2G | **Compare solutions to arrive at the most suitable one** | *Example: I am developing software and there are multiple solutions for a problem. I provide the AI with my use case, scale, and constraints, and **ask the AI to** analyze the pros and cons of each solution, then recommend the most effective one for my needs.*<br><br>*Example: I ask an LLM to rank the best potential solutions that I propose depending upon what the use case is, how I want end end product to look like, and user persona.* | I ask AI to create **a list of variations** that solve the problem in my stated thought process, and **explicitly request a pros and cons summary** of each solution.<br><br>The human provides AI with coalesced versions of potential solutions to get feedback on, that solve the problem. | AI writes code that ideally solves the problem with a summary of pros and cons.<br><br>The AI then based on the contextual knowledge of the system, evaluates the solutions, and ranks the solutions along with their pros and cons | I **evaluate each solution** using an IDE to determine whether each solution is **viable solution to the stated goal in** the thought process.<br><br>The human then evaluates this response for a sanity check, and then selects one system that is most vetted, most efficient, cost effective and has better chance of success and profitability. | - Determine trade-offs<br>- Scope solutions against constraints | - Programming fundamentals<br>- System architecture<br>- Tech stack<br>- Potential solutions | - Adaptable | - LLM<br>- other colleagues<br>- test environment |
| 2H | **Assess and optimize system resources for deployment planning** | *Example: I want to build a system with a certain large scale. I describe the product **then ask AI** to tell me what technical requirements the system should have given I will have a product of this particular size.* | The human should provide AI with what the **scale parameters** are, in terms of **how large the system** is in terms of **ability to process inputs at** a certain rate or size, the **budget** available, and **regions** where this system is deployed. | AI gathers this information and **suggests what is the hardware sizing of the system** in terms of **compute** and **storage**, and **what deployment aspects of scaling is needed** to ensure the system is deployed in a manner that is consistent with the ask | Human then evaluates this response for a sanity check, and then applies this suggestion in a POC basis to find if the suggestion is valid, and then documents the plan of action and the technical requirements in terms of actual hardware, and design considerations to get the implementation process started. | - Describe how information flows through the system and/or how different components of the system interact. | - System architecture<br>- Tech stack<br>- Various architectures | | - LLM based search tools (eg: web based search)<br>- AI integrated documentation tools |

# 3. Locate information (e.g., documentation, codelabs, API examples)

**Note:** For each of the tasks in this section, the cross-cutting SKAs from the previous section apply, in addition to the SKAs specific to each task.

## 03. Locate information (e.g., documentation, codelabs, API examples)

**Note:** For each of the tasks in this section, the cross-cutting SKAs from the previous section apply, in addition to the SKAs specific to each task.

| | Tasks | Examples | What the human does | What the AI does | What the human does | Skills | Knowledge | Attributes | Tools |
|---|---|---|---|---|---|---|---|---|---|
| 2A | **Research the latest trends, knowledge etc. to define the technical requirement that delivers the most compelling value proposition** | *Example: Given a topic or a problem, I ask an LLM to find article citations or research that happened in this area that would guide you to create your product.* | The human should provide AI with the **product spec**, the **performance** characteristics and the **input-output** characteristics of the system | AI then **breaks down the system into components**, and gathers information, literature and scholarly articles **about possible technologies and components**, and **research and advances in each and every component** that forms the backbone of the system. This research is provided in terms of scholarly articles, citations, conference talks and presentations about the **current state of the art** of each of every component in the system | Human, after **evaluating** the response **for sanity**, **documents the output** into forms of **competitor pitch**. This gives the human the delivery highlights of the system and the components that have to be hardened so much to tilt the customer attention in the product's favor | - Connect lower level technical concepts to the implementation of higher level application (and vice versa) | - Literature / Scholarly articles on Machine Learning.<br>- Literature / Scholarly articles on Software Engineering. | | - LLM based search tools (eg: web based search) |
| 2B | **Identify relevant open source libraries and documentation that can solve the problem at hand** | *Example: Given a topic or a problem, I ask an LLM to identify relevant open source libraries and documentation so the developer can formalize the final product.* | The human provides either the full or part of the **technical requirements** set and prompts the AI to **find relevant openly available libraries** that can be **useful for solving the problem** at hand | The AI processes the requirements provided, gathers information about existing openly available libraries, and **filters down** to the relevant libraries. The AI provides a **summary** of these libraries and **their relevance to the requirements outlined.** | The human, after evaluating the response, can then ask the AI to provide to expand on specific libraries and provide documentation on their usage | - Connect lower level technical concepts to the implementation of higher level application (and vice versa) | - Literature on industry trends and best practices (technical aspects).<br>- Literature on Open Source libraries. | | - LLM based search tools (eg: web search, internal documentation search) |
| 2C | **Create a GenAI "instance" as a domain expert** | *Example: Prompt an LLM to become a domain expert for internal and external knowledge consultation. For example, help you understand the code or produce new code.* | The human prompts the AI to play the role of a domain expert in a very specific field by **providing detailed information** regarding **a domain specific role and expectations.** | The AI generates domain-specific responses to any questions the users asks | The human uses the response and iterates to get a better understanding of new areas, including but not limited to unfamiliar code. | - Connect lower-level technical concepts to higher-level technical applications | - Types of domain experts to draw from | - Humble<br>- Observant<br>- Open-Minded | - LLM |

# 04. Generate insights from data

**Note:** For each of the tasks in this section, the cross-cutting SKAs from the previous section apply, in addition to the SKAs specific to each task.

# 04. Generate insights from data

**Note:** For each of the tasks in this section, the cross-cutting SKAs from the previous section apply, in addition to the SKAs specific to each task.

| | Tasks | Examples | What the human does | What the AI does | What the human does | Skills | Knowledge | Attributes | Tools |
|---|---|---|---|---|---|---|---|---|---|
| 4A | **Translate product (and technical) requirements into business needs (and vice versa)** | *Example: In all cases it's necessary to **get stakeholder buy-in** for product development. I ask an LLM to **"translate" and reframe product and technical descriptions into the framework of financial, business, strategic, etc. needs.***<br><br>*[additional comments received: "we usually start with a business need first and then model the changes to our products in order to serve that need."]* | *Human works with other teams/contributors in order to determine key product/technical requirements.*<br><br>*Human serializes these requirements in a format most understandable by the respective AI in their organization system.* | *AI, using both information provided by the human as well as past information, generates a mapping/translation between the business needs and the existing product capabilities + additional capabilities required to meet the needs.* | *Humans review the mapping generated by the AI system and use it to inform their planning and design decisions.* | - Re-evaluate pre-existing data model or mental model/perspectives about the specific business domain<br>- Describe the motivations of stakeholders<br>- Scope solutions to current constraints | - System architecture | | - LLMs<br>- other colleagues |
| 4B | **Determine the metrics to track** | *Example: I need to **track** business, product, and technical requirements and **transform them into relevant metrics to track success**. I ask an LLM to **read documentation and figure out the explicit and implicit needs / goals.*** | Human provides the **requirement** and **design documentation** as well as relevant **service codes** and **logs** to **AI** for it to **determine the appropriate metrics to track** based on **implicit and explicit goals**. | AI **produces a list of metrics** to track over time. | Human **creates systems and infrastructure to monitor** these metrics. | | - System architecture<br>- Tech stack<br>- Statistics<br>- Time-series analysis | - Observant | - Data analytics/monitoring platforms<br>- Instrumentation tools<br>- LLMs |
| 4C | **Extract pertinent subset of raw data** | *Example: One of the necessary steps in a data pipeline is **determining which data needs to be processed**. This becomes intractable for a person (or a team) to do when datasets contain 100s, 1000s, or 10000s of factors. I ask an LLM and AI tools in IDEs (integrated development environment) to determine which subset of data is needed by **either analyzing the code and data structures, or by being given the problem domain.*** | *The human gives the AI the dataset and provides the data requirement as well as analysis goals* | *The AI suggests data preprocessing and cleaning options. Once decided, it creates code to do the preprocessing (should be deterministic). Additionally, it can provide visualization of the dataset* | *Human decides which data transformations should be done and further processes the data* | - Visualize data flows | - Programming fundamentals<br>- Tech stack<br>- Ethics | | - LLMs<br>- IDE |
| 4D | **Transform unstructured data into structured data** | *Example: Raw data can be unstructured, semi-structured, or inconsistent and it's necessary to get into a structured format. It's much more efficient to **upload a sample of the raw data into an LLM and then describe the desired output in plain language**. I ask an LLM to **transform the data** and also **share the code** it used to perform the transformation.* | Human can provide **data dumps** to the AI and describe the **expected output format** based on their goals.<br><br>Human can also prompt AI to come up with **code** for this conversion so that it can be **reused later**. | AI either **converts** the data directly or **produces code** to perform this conversion. | Human verifies the converted data (or runs the code to perform the conversion) | - Work with various storage data formats | - System architecture<br>- Tech stack<br>- Data science | | - LLMs<br>- Data storage tools<br>- IDE |

## 04. Generate insights from data

**Note:** For each of the tasks in this section, the cross-cutting SKAs from the previous section apply, in addition to the SKAs specific to each task.

| | Tasks | Examples | What the human does | What the AI does | What the human does | Skills | Knowledge | Attributes | Tools |
|---|---|---|---|---|---|---|---|---|---|
| 4E | **Monitor metrics post deployment (e.g., performance issues, pagerduty alerts on certain predefined thresholds)** | *Example: After releasing a deployment, it's necessary to* ***monitor metrics for important events and patterns*** *(good and bad). I use AI tools - both LLMs and non-LLM-based tools - to* ***automatically monitor for relevant events****. There are typically a few methods that include rule-based thresholds, statistical outlier detection, and time-series analysis (for recurring patterns).*<br><br>*[additional comments received: "LLMs seem to do OK with time series data. For cases with very large amounts of data we allow them to write Python code to help analysis."]* | (for onboarding new systems for monitoring, and then next two are AI performing much of the detection/triaging while the human acts as the escalation contact)<br><br>Human sets up monitoring for new systems/services that is tied into the AI system.<br><br>Human ensures that the data pipeline properly represents the state of the system and that it is in a format that most compatible with the AI system (this is a bit general, and there's probably a bit of data hygiene role here that can be automated as well) | AI system continuously monitors the system/service metrics, identifies any performance issues or anomalous metrics, and triages them according to severity of issue as well as current/future human capacity.<br><br>For new issues the AI system is able to create a detailed trail of past, current, and expected future actions. A remediation strategy can be generated based on trained domain knowledge.<br><br>AI system brings a human into the loop for issues that are flagged and triaged via a centralized human alerting system (e.g. bug assignment, pager call etc.) The human is provided with the information they might need to remediate. | Humans respond to the issues raised by the AI. Humans can ask AI for more information or can provide detail on remediation strategy/progress. | | - Statistics<br>- Math<br>- Time-series analysis<br>- System Health Monitoring Tools | - Observant | - LLMs<br>- Cloud platform tools<br>- Time series analysis tools<br>- Statistical models |
| 4F | **Identify anomalous data events** | *Example: [from advisory board] I ask LLM to generate SQL query to explore logs so that I can identify unexpected anomalies with resolution.* | Human can provide data dumps or collected metrics from the running systems and ask the AI to **determine anomalous events** (e.g. differences in server logs that may be relevant). | AI identifies the events and can narrow down the time period on when it occurred. | Human reviews the validity of these results and take actions to fix these anomalies. | | - System architecture<br>- Statistics<br>- Math<br>- Time-series analysis<br>- System Health Monitoring Tools | - Observant | - Cloud platform tools<br>- Time series analysis tools<br>- Statistical models<br>- LLMs |
| 4G | **Present possible causes of observed data** | *Example: I prompt AI to* ***determine or infer the possible causes*** *for what we're seeing in the observed data, and AI can* ***return its hypotheses*** *about the underlying causes.* | Human finds some errors in the logs and identifies the related code, but isn't sure how it works | AI suggests cases that would cause those errors in the context of this system | Human tries to fix the identified issues; iterates. | | - System architecture<br>- Programming fundamentals | | - LLMs<br>- Root cause analysis tools<br>- Statistical models<br>- Logging tools |
| 4H | **Predict the results of potential actions to determine the best way to address observed anomalies** | *Example: once you have an idea of the possible causes for anomalies in observed data, prompt AI to* ***simulate new outcomes based on some possible set of actions****. For example, if AI infers that the underlying issue is CPU load, then it hints what the outcome would look like if I tweaked the CPU load.*<br><br>*[additional comments received: "if the system that is being changed is reasonably complex, simulating it becomes a Very Hard problem, and you should probably just make the change and test it with the real system."]* | Based on the **understanding of the anomalies**, human proposes **potential actions on how the problem can be fixed**. These actions can pertain to tuning the system settings and configurations. | AI simulates the results and can provide additional information on **what proposed action will work and what will not.** | Human looks at the results and **determines what action to take finally**. | | - Math<br>- System architecture<br>- System Health Monitoring Tools<br>- Simulation | - Observant | - A/B testing tools<br>- Multiple deployment environments<br>- Chaos engineering tools<br>- Simulation tools |

# 04. Generate insights from data

**Note:** For each of the tasks in this section, the cross-cutting SKAs from the previous section apply, in addition to the SKAs specific to each task.

| | Tasks | Examples | What the human does | What the AI does | What the human does | Skills | Knowledge | Attributes | Tools |
|---|---|---|---|---|---|---|---|---|---|
| 4I | **Generate reports that make insights interpretable by humans (visualized, clear)** | *Example: I need to communicate to stakeholders and I ask AI to **phrase what was done** and the **key results** in **less technical terms.*** | Human provides a list of changes and their impacts on metrics asks AI to summarize and visualize the data for non-technical stakeholder. | The AI generates an executive summary and charts to illustrate the key points. | Human reviews the report, shares with stakeholders. | | - Vocabulary (Buzzwords / Jargon) | - Observant | - LLMs<br>- LMMs |

# 05. Generate (create and maintain) holistic test coverage

**Note:** For each of the tasks in this section, the cross-cutting SKAs from the previous section apply, in addition to the SKAs specific to each task.

## 05. Generate (create and maintain) holistic test coverage

**Note:** For each of the tasks in this section, the cross-cutting SKAs from the previous section apply, in addition to the SKAs specific to each task.

| | Tasks | Examples | What the human does | What the AI does | What the human does | Skills | Knowledge | Attributes | Tools |
|---|---|---|---|---|---|---|---|---|---|
| 5A | **Identify all components (e.g., classes, modules, libraries) to test** | *Example: I have a big file and I prompt **AI to identify the components and generate a test plan**. Some code requires unit testing and others that require function testing.*<br><br>*[additional comments received: "There are various test types as acknowledged here, unit-tests, and functional tests, and the approach to these is quite different when working with AI. AI can usually give unit-tests as code, but functional tests are typically harder, since they involve a lot more components to be evaluated.]* | Human **provides access to the code** and coverage report and prompt to **identify missing test cases**. This could be directly integrated in the IDE or an external webpage. We are trying to **provide as much detail about the code that we want to test**. | AI looks at the existing test cases and its coverage and replies back with the types of tests to be written and why they matter. It gives a comprehensive view of the system's status in terms of testing. | Human looks at the suggestions from the AI and evaluates for importance and how to test it (e.g., testing plan). They can ask follow up questions for both of their tasks as well to get a better understanding of system. | | - Programming (e.g., how should the library code be implemented such that testing is easier and encapsulated)<br>- Testing Methodologies (e.g. what kinds of tests are needed, testing for API not implementation etc.) | - Organized | - IDE integrated with LLMs |
| 5B | **Identify potential points of failure within the scoped requirements** | *Example: I write a phone number parser. **I ask AI to break the code** (i.e., violate expected behavior). It **suggests a test that uses international numbers**. It also suggests **ways to fix** the code. I ask it to break the code again and it suggests **phonetic numbers** (Call 1800 JUNK) and suggests ways to fix the code to ensure that it works.* | Human provides the code to LLM and asks the LLM to **suggest inputs that would break the given piece of code**. | AI looks at the code and **generates various inputs** that would break the normal working of the code. This could be with lot of data, different data types, etc | Human looks at various scenarios generated by the AI and **evaluates** them on the **usefulness and adherence to the scoped requirements**.<br><br>*[comments received re test cases and scenarios: It all depends on what exactly we asked AI. E.g. sometimes I ask it to "suggest cases that should be tested", and it gives an English description of edge cases one would normally test (e.g. all 0s in a phone number). Other times, I directly ask it to generate test-cases, i.e. methods/ classes to test some behavior. So I would probably use "scenarios" here, because test-cases is limiting it to a very specific task.]* | - Evaluate test cases w.r.t usefulness and adherence to requirements | - Programming<br>- Testing frameworks<br>- Basics of unit and functional testing<br>- Code coverage analysis<br>- System architecture<br>- Programming Fundamentals | - Adaptable | - IDE integrated with LLMs<br>- Testing tools |
| 5C | **Test for base cases (e.g., ensures inputs & outputs of functions and programs are in the right data formats)** | *Example: I use AI to **try random tests** such as putting a number value for a text value.*<br><br>*Example: I have code that works with numbers as input data format. I ask AI to try to break the code. It suggests a failing test that uses string as input. It also suggests a typecheck to be added to the code to prevent it from breaking. It also suggests adding type hints for the precision of the outputs.* | Human **provides the code** and prompts **AI to write test cases** to test for input/output correctness | AI would suggest test cases which are of the same data type, different data types and try to break the code. | Human looks at the provided test scenarios and prioritizes the ones that they should fix | - Prioritize test cases to fix | - Programming<br>- Project requirements / functionalities<br>- Testing Methodologies<br>- Testing frameworks<br>- Programming fundamentals<br>- Tech stack | - Organized<br>- Accountable<br>- Observant | - IDE integrated with LLMs<br>- Testing tools<br>- Fuzzy testing tools integrated with LLMs |
| 5D | **Test for edge cases** | *Example: I have code that generates diff between two sentences. **I ask it to check for edge cases**. It suggests a test where the code doesn't work well when trying to generate a diff between 1st and first. It also suggests fixes for the code to make the test work.* | Human provides the code to LLM and asks the LLM to suggest inputs that would break the given piece of code. | AI looks at the code and generates various inputs that would break the normal working of the code. This could be with lot of data, different data types, etc | Human looks at various scenarios generated by the AI and evaluates them on the usefulness and adherence to the scoped requirements. | | - Programming<br>- Project requirements / functionalities<br>- Testing frameworks | - Organized | - IDE integrated with LLMs<br>- Testing tools |

# 05. Generate (create and maintain) holistic test coverage

**Note:** For each of the tasks in this section, the cross-cutting SKAs from the previous section apply, in addition to the SKAs specific to each task.

| | Tasks | Examples | What the human does | What the AI does | What the human does | Skills | Knowledge | Attributes | Tools |
|---|---|---|---|---|---|---|---|---|---|
| 5E | **Ensure tests remain organized and consistent after the code to be tested has been refactored** | *Example: I refactor a function out of file A into a new file B. (I ask AI to **move the test code to the appropriate file and fix the references**) Similarly, if I split a complex function into two, I ask AI to take care of splitting the tests into separate files as well.* | Human provides before/after for non-test code to the AI, and prompts for refactoring of the respective tests. | AI would suggest a refactoring to align the tests with non-test code. | Human looks at the suggestion, and accepts the refactoring after evaluation.<br><br>If the refactoring is partial, human works on the remaining part to complete the refactor. | - Organize code<br>- Write well factored code | - Testing methodologies<br>- Testing frameworks<br>- Software best practices | - Adaptable<br>- Empathetic<br>- Organized | - IDE integrated with LLMs |
| 5F | **Generate a sample of production test data to imitate real-world use cases** | *Example: I have an API endpoint and I want to test it with a sample representative of what a production user would do. I ask AI to **generate a few samples for each use case that I have in mind**. It generates test cases that look similar to what a user in production would be using my API for.* | Human provides AI access to realistic production data | AI analyzes production data, learns from it, and responds back with a sample of synthetic data similar to production data, to be used in tests. | Human evaluates the characteristics of the generated data, compares it with the characteristics of production data, and determines if and which parts to use for testing. | | - Production pitfalls | | - LLMs<br>- IDEs integrated with LLMs |
| 5G | **Test performance with large volumes of data** | *Example: I have an API endpoint and I want to test how it will **perform under load**. I ask AI to generate **load testing code with real-world examples**. It generates the code to test the API in more real life like methods, rather than approximations of real use cases.* | Humans provides the code and asks for how to load test that code | AI provides guidance on simulating to send large amounts of data wither from scratch or using existing tools<br><br>AI **writes code** that can be used **to simulate large amounts of data** to the API or function. | Human runs these tests (manually or through tooling) to assess system performance.<br><br>Human runs these tests to gather what is working and breaking | | | | - LLMs<br>- IDEs integrated with LLMs<br>- Load testing tools integrated with LLMs |

# 06. Investigate issues in production

**Note:** For each of the tasks in this section, the cross-cutting SKAs from the previous section apply, in addition to the SKAs specific to each task.

# 06. Investigate issues in production

**Note:** For each of the tasks in this section, the cross-cutting SKAs from the previous section apply, in addition to the SKAs specific to each task.

| | Tasks | Examples | What the human does | What the AI does | What the human does | Skills | Knowledge | Attributes | Tools |
|---|---|---|---|---|---|---|---|---|---|
| 6A | **Triage active production issues** | *Example: I have an integrated system connected to an AI model, and I ask AI to **recommend how to prioritize the active production issues.*** | Human(s) create(s) a prioritized list of critical user journeys, and provides this along with a list of bug reports to the AI, and asks it what bugs should be prioritized. | The AI prioritizes the list of bugs according to which CUJ they occur in and how severe the issue is (brokenness per user, and total number of users affected) | Human checks the response and starts working on the top bugs. | - Plan project<br>- Manage time<br>- Prioritize<br>- Determine trade-offs<br>- Integrate human + AI knowledge | - Tech stack<br>- System architecture<br>- Security flaws<br>- Design patterns<br>- Programming fundamentals<br>- Product architecture | - Efficient<br>- Observant<br>- Transparent<br>- Systematic | - Logging systems<br>- Event monitoring<br>- AI analysis tools<br>- System dashboards |
| 6B | **Replicate and revisit issues (in a controlled environment) (e.g., tests) to evaluate fixes and future proof** | *Example: Given the error logs and the associated test cases, I ask AI to **identify a new test case** which could have caught this issue earlier.*<br><br>*Example: Once I know how to generate the error, I use AI to fix bugs by **finding how to recreate the issue and test if the error happens again**.* | The human must prompt the AI to **investigate why an error is not being caught** / handled properly.<br><br>The human must provide the **case**, the **error**, and the **handling code**. | AI **identifies the logical mistake** or **configuration** issue that might have caused the problem. It also **utilizes up to date information to suggest accurate and up to date responses**. | The human now needs to evaluate AI's response.<br><br>Human **applies the suggested configuration and code changes in a sandbox to validate the suggestions**.<br><br>On successful identification of the core issue in sandbox, the change is ready for application in the production system. | - Grasp how configurations affect the systems<br>- Identify base cases and edge cases as writing / from reading code<br>- Predict ways the application is used and could be used/abused | - Testing methodologies<br>- Testing frameworks<br>- Understanding of configuration management systems<br>- Codebase<br>- Systems and sandboxing processes | - IDE<br>- LLM<br>- Testing tools<br>- Configuration management tool<br>- Resource management systems | |
| 6C | **Determine source of issues conceptually by reviewing charts (anomalous events and observed data) and summarizing logs and traces** | *Example: There's an error or possible issue in production. I provide AI with error logs and stack traces and ask AI to analyze and suggest areas to investigate. It analyzes and suggests possible areas for investigation.*<br><br>*Example: Given the profiling data I ask AI to identify issues that are not apparent but might cause issues under load.*<br><br>*Example: I prompt AI to determine or infer the possible causes for what we're seeing in the observed data, and AI can return its hypotheses about the underlying causes.* | Right now, humans have to **provide AI with all the context** including **systems** where the error happened, **the error log,** the **source code** and any other **system architecture information**.<br><br>There are newer tools coming in which are **integrating** some of these systems together so that humans don't have to copy paste.<br><br>The human must prompt the AI with **historical stack traces**, **system logs** and even **user outputs** to identify what might be going on with the system to cause higher latency or inaccurate responses. | AI identifies the error and utilizes the provided source code and documents as context to iteratively suggests areas to dig deeper into. Presents the user back with details on what might have gone wrong.<br><br>AI identifies **trends** in the system and thus helps pinpoint the problem in the code or the system. | The human now needs to evaluate AI's response for **sanity check**.<br><br>The human may need to provide more documents and context to get a better answer and for the AI to **pinpoint** source of issue and the solution. | | - Project functionalities/ requirements<br>- Distinguishing production errors vs. code bugs.<br>- The kinds of things that go wrong in production<br>- Where to find the logs and traces.<br>- Where to find the config files<br>- How to run a code diffs<br>- Systems involved in production deployment<br>- Code tracing frameworks<br>- End to end systems<br>- System architecture<br>- Programming (fundamentals) | - Level-headed<br>- Observant | - IDE<br>- LLM<br>- Error capturing and tracing tools<br>- Resource monitoring tools<br>- Root cause analysis tools<br>- Statistical models<br>- Logging tools |

# 06. Investigate issues in production

**Note:** For each of the tasks in this section, the cross-cutting SKAs from the previous section apply, in addition to the SKAs specific to each task.

| | Tasks | Examples | What the human does | What the AI does | What the human does | Skills | Knowledge | Attributes | Tools |
|---|---|---|---|---|---|---|---|---|---|
| 6D | **Isolate suspected broken code and configurations** | *Example: Based on the error logs and code, I ask AI to **pinpoint where** in the code **the issue is** coming from.* | The human has to iteratively work from a generic production issue to a more specific problem in the code or configuration.<br><br>This involves multi-turn conversations with the AI and providing it with a lot of documents and details. | AI iteratively suggests areas to dig deeper into and isolate the code that might be the cause of the error. if it has access to code and config, often **pinpoints the exact root cause** for the production issue. | The human needs to evaluate AI's responses and ask more questions to keep digging deeper.<br><br>Often the production problems are less about code and more about configuration. So humans have to apply the config and monitor the changes.<br><br>The human identifies what needs to change and applies the appropriate fixes to the system and code. | - Grasp code and systems to better guide the AI to stay on correct course of action | - Project functionalities<br>- Systems involved in production deployment<br>- Codebase<br>- Configuration management systems | | - IDE<br>- LLM<br>- Configuration management system<br>- Resource monitoring tools |
| 6E | **Apply changes and evaluates fixes** | *Example: I have an integrated system connected to an AI model, and I ask AI to recommend how to address them.* | Gives AI the error logs, trace logs or whatever alerted the team to a possible production issue | AI gives a possible resolution of the issue and asks for more questions and suggest various areas to look into to fix the issues | Human looks the relevant code, metrics and other details to identify what to update and iterates with AI until the issue is resolved.<br><br>The human identifies what needs to change and applies the appropriate fixes to the system and code. | | - Tech stack<br>- System architecture<br>- Security flaws<br>- Design patterns<br>- Programming fundamentals<br>- Product architecture | | - IDE<br>- Multiple test environments |
| 6F | **Explore potential remediations to prevent similar production issues in the future** | *Example: I provide AI with all the events, error logs, system diagnosis and then **I ask AI** to generate retrospective which includes remediation items.*<br><br>*[additional comments regarding the task: "optional and probably not done for the majority of bug fixes"]* | The human must provide all the **logs**, **traces** and **system information** to the AI. The human also must provide a **template** to be used **for the retrospective**. | AI **summarizes various actions** that were taken to **identify** the root cause and actions taken to **apply** the fixes. It also **suggests remediation items**. | The human **applies the remediations** suggested by the AI. | - Project estimation | - Programming<br>- Project functionalities<br>- Codebase | | - IDE<br>- LLM<br>- Testing tools<br>- Resource management system |
| 6G | **Write retrospective document / report** | *Example: If something went wrong, **use LLM** to create a retrospective / post-mortem report to share with stakeholders*<br><br>*[additional comments regarding the task: "postmortem document" might be done for high severity issues only depending on the organization]* | Human finds all of the relevant artifacts related to the outage - emails, chat logs, bug reports, metrics, server logs... | AI generates a postmortem based on these | Human reviews and socializes wtih the PM | - Integrate human + AI knowledge<br>- Technical writing | - Tech stack<br>- System architecture | - Honest<br>- Perfectionist<br>- Transparent | - LLMs<br>- documentation tools<br>- collaboration tools<br>- knowledge management tools |

# 07. Contextualize the work item

**Note:** For each of the tasks in this section, the cross-cutting SKAs from the previous section apply, in addition to the SKAs specific to each task.

# 07. Contextualize the work item

**Note:** For each of the tasks in this section, the cross-cutting SKAs from the previous section apply, in addition to the SKAs specific to each task.

| | Tasks | Examples | What the human does | What the AI does | What the human does | Skills | Knowledge | Attributes | Tools |
|---|---|---|---|---|---|---|---|---|---|
| 7A | **Establish user persona (e.g., user challenges, pain points, and needs)** | *Example: Ask AI to inform the context of the solution of the user persona **created from commonalities** among individuals requesting feature/reporting bug.*<br><br>*Example: I ask the AI to create a user persona.*<br><br>*Example: Once a persona has been established by the developer team, I give the specifics about this persona to the AI and ask it to tell me about any missing gaps that were missed out about that particular user persona (i.e. find more information about the persona).* | The human uses the contextualized feedback, **provides the relevant users profiles** as context and prompts the AI to understand **commonalities** between various users<br><br>The human first provides any context about the problem space and potential application/product in question. Using the context, the human prompts the AI to **provide potential user personas, pain points** faced by these users, and **features** that the user needs. | The AI **generates a common user persona** that are impacted by the specific work item<br><br>The AI **processes** the context, gathers information about various users and provides details and a **breakdown of the potential market, user types, and user needs.** | The human uses this user persona to better contextualize the work item and better understand the type of user impacted by the solution<br><br>The human evaluates the different users suggested and performs a sanity check to match with known intuition. The human then focuses on specific user personas and iterates with the AI to further flesh out more detailed user needs. This information will help **identify the right application space and inform technical requirements** | | - Domain knowledge of user-facing application and tools<br>- User characteristics<br>- Common usage characteristics of the product | - Empathetic | - LLM<br>- Modeling tools |
| 7B | **Analyze feedback from users and internal stakeholders** | *Example: Ask AI to analyze a large data set to find commonalities among feedback from users/ stakeholders*<br><br>*Example: I am receiving feedback from various sources and I use AI to **know where the feedback came from** and to **filter** the feedback down **to a need I can solve for**.* | The human gives the AI the context for the desired output with a prompt such as, "From the following list of feedback, please group by common themes and provide these common themes in order of frequency along with and any potential analyses of the sources of that theme." The user then provides the list of feedback with any metadata about the source/author/ | The AI provides a list of common themes from the feedback that it has grouped together in order of priority, along with any pertinent information. | The human reviews this and cross checks with a sample of the feedback to ensure accuracy and the correct level of generalization. They may prompt for more detail or more items in the list of grouped themes. | - Obtain or integrate data from multiple sources to feed them to AI | - Data formats<br>- Programming fundamentals | | - LLM |
| 7C | **Integrate needs and feedback into work item** | *Example: Ask AI to connect and correlate the user feedback and internal stakeholder feedback on a work item.* | The human provides internal and/or external stakeholder feedback as context. Along with additional context about the specific task, the human prompts the AI to filter down and summarize relevant feedback pertaining to the task at hand | The AI retrieves the appropriate pieces of feedback from the large corpus of feedback and summarizes them for the human to review. The AI also provides explanation for how it is connected to the specific work item | After verification, the human can then use this information to gain a better understanding of how the task impacts internal and external stakeholders | | - Domain knowledge of user-facing application and tools | | - LLM |

# 07. Contextualize the work item

**Note:** For each of the tasks in this section, the cross-cutting SKAs from the previous section apply, in addition to the SKAs specific to each task.

| | Tasks | Examples | What the human does | What the AI does | What the human does | Skills | Knowledge | Attributes | Tools |
|---|---|---|---|---|---|---|---|---|---|
| 7D | **Investigate competitive solution available in the public domain** | *Example: **I ask the AI** to tell me the technical requirements of my product. If there are any competitors working on the same problem AI can express how competitors are solving the problem. With that information, **I ask the AI** to tell me how to make the product even better or more competitive than the current competitors in the market.* | The human must tell the AI the product behavior envelope. Its output styles, its input styles, the expected outcome and the differentiator. He/She then requests AI to suggest similar products in the market | The AI then gathers the information, and suggests competitive products available in the market. Assuming the AI is trained on the review comments for the products, AI then suggests **most hated and most loved feature sets** of these competing products, and suggests what can be **done within the budget to make the product better than competition** | Human then evaluates this response for a sanity check, and then documents the suggestions thereby **constituting a well formed acceptance criteria** for the product that, at least on paper is highly desirable and profitable, than its competitors in the market | - Search, and retrieve current solutions in the market to provide to the AI. | - Industry trends (both technical and business) | - Observant | - LLM based search tools (eg: web based search)<br>- Tools that provide access to market trends and analysis |
| 7E | **Extract scope and assumptions for technical requirements and decisions** | *Example: Ask AI to **summarize** and **highlight the key points** of the scope and the assumptions that would be crucial to technical implementation decisions* | The human initially provides **context** for the AI's mindset and response, reminding it **not to miss any detail that could be crucial to implementation decisions** but to **condense/distill** as much as possible the following content and referenced bodies of content. The AI is then furnished with the unabridged content. | The AI reduces the content to its **key points** and **outputs a summary**. | The human consumes the summary and further prompts for more information on each point as necessary. | | - Limitations of existing application stack | | - AI integrated documentation tools<br>- LLM |
| 7F | **Refine requirements** | *Example: Given the technical requirements, user persona, and solutions in the public domain, I ask an LLM to produce a final set of requirements.* | The human provides context obtained as a result of the previous steps - technical requirements, user personas, existing competitive solutions, and asks prompts the AI to **refine** this into a final requirements set | The AI uses the collected information to **consolidate ideas** across different steps, remove any redundancies, and provide a final refined set of technical requirements | The human then **evaluates the consistency of the requirement document** and can iterate further with the AI to correct/refine specific parts | - Identify and question underlying assumptions | - System architecture<br>- Tech stack | | - Requirement comparison and analysis tools<br>- Requirement and pricing correlation tools |
| 7G | **Connect task at hand to broader context of the larger software system** | *Example: I am a programmer who is new to this codebase and I need to understand the high level of what the code is doing. I ask AI to **give me a synopsis of what I am working on** and **how my task fits into the broader codebase and system**.* | The human provides the **broader codebase** and any available **documentation** as context. Along with additional context about the **specific task**, the human prompts the AI to summarize where the task fits into the larger application/software | The AI produces a **summary of how the task is connected** to the broader work item. Ideally, the AI **provides sources and pointers to the relevant parts of the codebase** | The human can then **verify** the output of the AI and **iterate** with the context to go into **finer detail within specific parts of the codebase** | | - System components available, their capabilities and caveats<br>- Domain knowledge of user-facing application and tools | - Observant | - LLM<br>- LMM |

# 08. Ensure launch complies with legal, privacy & security requirements

**Note:** For each of the tasks in this section, the cross-cutting SKAs from the previous section apply, in addition to the SKAs specific to each task.

## 08. Ensure launch complies with legal, privacy & security requirements

**Note:** For each of the tasks in this section, the cross-cutting SKAs from the previous section apply, in addition to the SKAs specific to each task.

| | Tasks | Examples | What the human does | What the AI does | What the human does | Skills | Knowledge | Attributes | Tools |
|---|---|---|---|---|---|---|---|---|---|
| 8A | **Determine potential patent and/or copyright infringements based on product specification** | *Example: I need to check my code isn't infringing on issued patents, so I input my product to an AI model trained on patents and figure out what are the most similar looking in-profit patents that have been issued in the same area.* ***I ask AI to determine if my product is infringing on any patents or copyrights.*** | The human feeds the AI with the input. The input can be codebase, design specification, engineering requirements document or the functional envelope of the system with inputs and outcomes. Inputs can be all, many or few of the above categories. | The AI then sieves through its model, finding potential patent violations and copyright infringements along the functional spec of the current product. It also publishes a measure of similarity, which is a measure on how closely the current specification matches with an existing patent or copyright | The human, after reviewing the output for sanity, uses this output and reviews the measure of similarity for various artifacts within the product spec and makes changes as appropriately to ensure copyrights and patents are not infringed upon | | - Tech stack<br>- Industry trends | - Forgiving | - Patent search tools<br>- AI Enabled scholarly article and asset search, or legal search tools |
| 8B | **Examine dataflow and product footprint to detect privacy violations** | *Example:* ***Use AI*** *to detect privacy violations by integrating data flow logs and product usage details to define machine-readable privacy policies. Train anomaly detection models to spot deviations, monitor data flows and product footprints in real-time, and generate alerts for potential violations.*<br><br>*I need to be familiar with the General Data Protection Regulation (GDPR) to build an implementation that is legally compliant. I use AI to gain a basic understanding of the technical requirements before the lawyer colleague becomes available. Additionally, since ensuring GDPR adherence can be complex and error-prone, I use AI to consume all logs produced by a single user to test one specific user journey and suggest potential policy violations.* | The human feeds the AI with the input. The input can be codebase, design specification, engineering requirements document or the functional envelope of the system with inputs and outcomes. Inputs can be all, many or few of the above categories. | The AI then examines the functional spec, data movement, design aspects, storage aspects to ensure privacy or data violations are absent. For any potential violation that is discovered, it is duly flagged and reported to the human. | The human, after reviewing the output for sanity, uses the violations related data and gathers more information to form potential fixes to the privacy violations. These fixes are documented as compulsory fixes before the product is released into the market. | | - Perceived output of the components<br>- Product architecture<br>- System architecture, interdependencies between components | | - Data visualization tools<br>- Debuggers |
| 8C | **Generate penetration test cases and bad inputs to detect security violations** | *Example:* ***Use AI*** *to generate penetration test cases and malicious inputs to break the system. Automate these test cases based on identified vulnerabilities and attack patterns, employing adversarial learning techniques. Execute these tests in controlled environments to detect security violations such as unauthorized access and data breaches. Analyze results to prioritize vulnerabilities and produce detailed reports with mitigation strategies.* | The human feeds the AI with the input. The input can be codebase, design specification, engineering requirements document or the functional envelope of the system with inputs and outcomes, and the way the inputs are fed into the system and the authentication mechanisms. | The AI then generates penetration inputs, both valid and malicious, and input types and sizes, to potentially break the system. | The human then validates these inputs, and documents these penetration test cases in test automation frameworks to be automated by software test infrastructure, with onus on privacy and security violations | | - Entry points into the software product<br>- Tools | - Ethical-hacker Mindset | - Vulnerability scanning tools |
| 8D | **Attempt to penetrate the system with simulated attacks** | *Example:* ***Use AI*** *to attempt to penetrate the system by itself with simulated attacks. For example, a simulated attack could be an SQL injection, which AI could readily impose on the system or product that I am building.* | The human feeds the AI with the system characteristics and input envelope and the penetration test cases, and requests the AI to penetrate the system | The AI then penetrates the system in the provided ways documented by the penetration test cases, and captures the after effects of the system after the penetration attacks complete, and generates the post penetration state of the system after each penetrated simulated attack | The human studies the post attack characteristic of the system and verifies the expected and real outcome of the penetration attacks. If there is any anomaly detected in these attacks, such anomaly is studied and fixes are drafted into place to contain the damage after a real such penetration attack | | - Cybersecurity fundamentals<br>- Tech stack<br>- System architecture | - Courageous<br>- Ethical-hacker Mindset<br>- Observant | - Vulnerability scanning tools<br>- Vulnerability exploitation tools |

# 08. Ensure launch complies with legal, privacy & security requirements

**Note:** For each of the tasks in this section, the cross-cutting SKAs from the previous section apply, in addition to the SKAs specific to each task.

| | Tasks | Examples | What the human does | What the AI does | What the human does | Skills | Knowledge | Attributes | Tools |
|---|---|---|---|---|---|---|---|---|---|
| 8E | **List areas / components within the product needing security enhancements with examples** | *Example: **Prompt AI** to list components with flaws that I need to fix. It also gives me the inputs to reproduce the problem. For example, if storage within email is compromised, then storage needs rework. It can suggest exploring alternative cloud storage.* | The human points the AI to the codebase and API configurations for the product, as well as feeds of known security vulnerabilities in its dependencies, and findings from non-AI static analysis tooling. | The AI determines which findings have likely potential impact on the product, and ranks them according to security and privacy risk. | The human helps triage the ranked findings from the AI, ensuring remediation work is assigned to the appropriate owners. | - Determine tradeoffs / perform cost-benefit analysis | - Product architecture<br>- Programming fundamentals<br>- Tech stack<br>- Codebase | - Critical Thinker<br>- Organized | - Debugger<br>- Diagraming tools<br>- Knowledge mgmt tools<br>- Project mgmt tools |
| 8F | **Generate disclaimer notes detailing known flaws** | *Example: I need to protect my team from probable lawsuits for my beta product, and **I use AI to** produce text from context of my security flaws to alert the users in the terms of agreement.* | The human gives the AI a list of known security flaws. This list can also be curated by AI but the human should have the option to add/remove flaws | The AI drafts terms of agreement based on the list of flaws. It should act as a lawyer and can retrieve specific terms from a database. | The human helps triage the ranked findings from the AI, ensuring remediation work is assigned to the appropriate owners. | | - Industry knowledge<br>- Tech stack<br>- AI fundamentals<br>- Programming fundamentals | - Empathetic<br>- Observant<br>- Perfectionist | - Collaborative documents<br>- LLM |
| 8G | **Propose possible solutions to bring the product back to safety & compliance** | *Example: Given that AI has contextual knowledge until this far, **I ask AI** to propose possible solutions to bring the product that we are building back to safety and compliance standards by detecting problems and contributing possible solutions.* | The human feeds the AI with tickets describing open security/privacy issues in need of remediation, as well as technical details of the systems involved. | The AI proposes concrete remediation work to prioritize. This may be in the form of localized fixes to code or ACLs, or larger scale re-architecture. | The human reviews and approves proposed changes and determines appropriate next steps to get those changes into production. | - Define a rollout strategy<br>- Determine trade-offs of possible solutions | - Product architecture<br>- Industry knowledge (compliance, legal)<br>- Tech stack | - Empathetic<br>- Ethical | - Collaboration tools<br>- Project mgmt tools<br>- Legal references |

# 09. Develop and document high-level technical designs

**Note:** For each of the tasks in this section, the cross-cutting SKAs from the previous section apply, in addition to the SKAs specific to each task.

# 09. Develop and document high-level technical designs

**Note:** For each of the tasks in this section, the cross-cutting SKAs from the previous section apply, in addition to the SKAs specific to each task.

| | Tasks | Examples | What the human does | What the AI does | What the human does | Skills | Knowledge | Attributes | Tools |
|---|---|---|---|---|---|---|---|---|---|
| 9A | **Determine goals and non-goals of project (e.g., technical scope and objectives of the system/feature being designed)** | *Example: After the information gathering, I need to determine what to work on, so I use AI to **cluster data from prior research** (e.g. log messages, user feedback, etc), to **finalize** and **refine goals**.* | Human provides **documents**, **conversations**, and other **research** and **relevant data** (e.g. log messages, user feedback, product requirements, design briefs, etc.) to the AI and asks the **AI to identify various goals** and **requirements** of the project. | AI **summarizes** the provided details and from within it **extracts specific asks.** It then sets them up in a "goals and non-goals" sections.<br><br>AI analyzes the provided documents to extract key information related to scope and objectives, identifies potential ambiguities or inconsistencies in the human's input, and generates a draft scope statement summarizing the problem, target users, and intended outcomes. | Humans look at the suggestions of goals and non-goals and iterates with the AI to get to a good understanding of the asks as well as more details on each section. | | - Industry knowledge | - Observant | - LLM integrated data analysis tool to find patterns<br>- LLM integrated writing tool for summarization |
| 9B | **Review AI suggestions of appropriate document templates** | *Example: While creating documents with specific section requirements, I ask AI **which template is appropriate** for the tasks either from internal company documents or external public information.*<br><br>*Example: "what's a great way to show how much more impactful using service X is over Y? Here are dates and latencies, I want to make it clear"* | Human provides context on the goal of the document and appropriate sections that may be needed in that doc. | AI **searches the internal company** documents or documents with a similar outline over the public web to **suggest a template suitable for the task** | Humans review the suggestions and iterates with the AI if needed. Finally an appropriate document template is picked. | | - General knowledge about technical documentation | - Efficient | - LLMs<br>- Document Template Repositories |
| 9C | **Expand on key ideas about the technical requirements and potential solutions** | *Example: I need to transform my ideas into code and I ask an LLM to take my ideas in the form of bullet points and **expand on them to produce more comprehensive details**.* | Human provides a **key idea and context of the problem domain** where that idea is to be applied. They ask the AI to use these details to generate more details.<br><br>Human provides an outline of the steps that they would like to take in order to solve a technical problem. | AI **contextualizes** the problem and **suggests** ways to apply the idea **with more details on how to apply**<br><br>AI uses all the context and the rough outline described by the human to generate a more detailed documentation and suggests improvements. | Human **iterates** with the AI to get a better understanding of **how to apply the key idea** in the correct context and then **adds these details to the documentation**<br><br>Human iterates with AI to come up with a better overall solution.<br><br>*[Context offered by participant - V] The way I like to think of this is the same way you would do a youtube video. You start with the outline of thing that you want to talk about, then iterate to get to a overall script that you would use.* | - Connect lower level technical concepts to higher level technical applications and vice versa | - Tech stack<br>- System architecture<br>- Software product design<br>- Product architecture<br>- Industry domain knowledge | - Efficient | - LLM integrated writing tool<br>- LLM integrated mind-mapping tool |

# 09. Develop and document high-level technical designs

**Note:** For each of the tasks in this section, the cross-cutting SKAs from the previous section apply, in addition to the SKAs specific to each task.

| | Tasks | Examples | What the human does | What the AI does | What the human does | Skills | Knowledge | Attributes | Tools |
|---|---|---|---|---|---|---|---|---|---|
| 9D | **Develop visuals (e.g. design diagrams) to communicate ideas** | *Example: Import text or sketched diagram to the AI model and ask it to generate a block diagram out of the input sketch or paragraph.* | Human provides a **hand-drawn sketch** of the design diagram or **a description** of the diagram in a textual format with appropriate inputs and outputs. | AI produces a **refined design diagram** from the inputs. | Human **reviews** the final diagram produced and **re-iterates** if needed. | - Visualize flow of information | - System architecture<br>- Design patterns<br>- Industry domain knowledge | - Efficient | |
| 9E | **Generate technical roadmap (i.e. bigger picture)** | *Example: Collaborate with an LLM to create a **plan** (technical decision already made) and **timeline** by detailing the steps that need to happen and the time it might take for each task.* | Human provides all the **unordered tasks-to-be-done** to the AI and asks AI to **prioritize** and **parallelize** based on the **available resources** and **product deliverables** | AI tries to **reason against the tasks and the required deliverables** to get to a good **priority** of the products and suggests a **timeline** of events and actions | Human iterates with AI by giving in additional constraints until all the requirements are satisfied. | - Determine tradeoffs<br>- Cost estimation | - System architecture<br>- Software product design<br>- System architecture<br>- Tech stack<br>- Skills of various team mates | - Observant<br>- Time Consistent | - LLM integrated planning tool<br>- Access to skills data for teammates |
| 9F | **Include references & dependencies (e.g., backlinking design doc to support technical design decisions)** | *Example: Use an LLM to create a documentation of the plan and choices taken.* | Human provides an initial document for which they want the **references** to be generated and the **format** in which they are needed. | AI accesses the internal company document repository and other public sources to generate references. | Humans verify the generated information. | | - Industry domain knowledge<br>- Related technical literature | | - LLMs<br>- LLM integrated Text Editors |
| 9G | **Condense relevant information (to record in the plan/doc) for comprehensibility** | *Example: Ask AI to **condense** the information so it communicates the relevant sections to each stakeholder for better comprehension and in a **digestible** fashion.* | Human provides **conversation logs, or a long discussion thread** to the AI and asks it to generate a more **digestible summary** | AI identifies the **key themes** and **topics** discussed and tries to get to an **actionable** summary | Human iterates with AI to get to a summary they understand and is useful for their task | | - Various data formats<br>- Data patterns<br>- Industry domain knowledge | | - LLM integrated review tool<br>- LLM integrated visualizer |
| 9H | **Incorporate feedback to various parts of the documentation** | *Example: I need to take in feedback from different stakeholders that may apply to various areas of the documentation the stakeholders are unfamiliar with, and I ask AI to point out **what changes** need to be made **in which sections** of the document that reference the changes made.* | Human ask the AI on how to incorporate the feedback received through the stakeholders in the document. | AI comes up with a revision of the document. | Human reviews all the AI suggestions, makes the best decision using his domain knowledge and creates the final version. | | - Tech stack<br>- Design patterns<br>- Product architecture | - Efficient | - LLMs<br>- LLM integrated Text Editors |

# 10. Generate assets

**Note:** For each of the tasks in this section, the cross-cutting SKAs from the previous section apply, in addition to the SKAs specific to each task.

## 10. Generate assets

**Note:** For each of the tasks in this section, the cross-cutting SKAs from the previous section apply, in addition to the SKAs specific to each task.

| | Tasks | Examples | What the human does | What the AI does | What the human does | Skills | Knowledge | Attributes | Tools |
|---|---|---|---|---|---|---|---|---|---|
| 10A | **Research styles and references** | *Example: **I prompt** an AI image generator to give examples of different tigers to obtain an exhaustive list of the tigers to see all styles and angles, positions, etc* | Human prompts the AI to generate example assets based on their needs | AI generates these examples | Human reviews the results and re-iterates with different prompts if needed. | - Scope the project<br>- Brainstorm and describe the desired output using text or other modalities (voice, reference image + text) | - Types of art | | - GenAI |
| 10B | **Optimize prompts using AI to produce assets** | *Example: **Ask AI** (e.g., Gemini, Mid Journey, etc.) to create a prompt for another AI model to respond to with assets.* | Human prompts an AI model to **create an appropriate prompt for another AI model based on its knowledge of this other model**. Human can also give examples of **what worked** and **what failed** when they manually tried prompting the other model. | AI generates prompts suitable for the other model. | Human uses the output prompt, makes modifications and tries it on the other model. Human re-iterates if needed based on the results. | - Train a pre-existing AI model to suit your use case/domain (e.g., by uploading spreadsheet of values) | - Limitations of the Tools<br>- Various data formats | | - GenAI<br>- LLMs |
| 10C | **Convert between modalities** | *Example: **Prompt AI** to produce an image from text or generate a text to caption an image.* | Human gives the inputs in a certain modality, describes the output expectations and its modality | AI produces the results in the modality needed | Human reviews the results and re-iterates if needed. | | - Tools for various modalities that are available and their limitations. | | - GenAI |

# 11. Improve reliability to avoid production problems

**Note:** For each of the tasks in this section, the cross-cutting SKAs from the previous section apply, in addition to the SKAs specific to each task.

## 11. Improve reliability to avoid production problems

**Note:** For each of the tasks in this section, the cross-cutting SKAs from the previous section apply, in addition to the SKAs specific to each task.

| | Tasks | Examples | What the human does | What the AI does | What the human does | Skills | Knowledge | Attributes | Tools |
|---|---|---|---|---|---|---|---|---|---|
| 11A | **Identify points of failure for critical components along system workflows and potential remediation items** | *Example: I give AI documentation or diagrams for the system or code itself and I ask AI to identify additional reliability considerations.*<br><br>*Example: I give AI a cloud architecture diagram and ask it to generate statements that instruct me to consider adding a failback database or a load balancer in another region to make sure there is some redundancy.* | Gives AI access to the **overall system architecture** and asks it to **identify critical components of the system that might fail** during high load or unforeseen events | AI assess these details to **identify** critical components and **might ask more questions to better understand the context** in which the architecture is running. It may give remediation items that can be used to optimize the infrastructure | Human **evaluates** these response and **iterates** with the model to get to a **better understanding** of the system components and either **apply** the suggested changes or **keep a note** of them in the failure guides. | - Evaluate severity of the scalability and reliability weaknesses given the current state of the world (resource costs, DoS attack risk, product reputation, planned launches). | - Tech stack<br>- System architecture<br>- Design patterns<br>- Security flaws | - Observant<br>- Perfectionist<br>- Transparent | - LLM integrated resource planning tool (on cloud provider)<br>- LLM based mind mapping tool |
| 11B | **Add logging and monitoring to critical system flows** | *Example: I am working on code that is large and complex and I need to add log messages where relevant. I know what might go wrong. I ask the AI system to identify critical paths and apply a log message to that area, including creating the log message for me that will print out information when the code is executed.* | Gives AI access to the **code** and prompts it to **suggest what information to capture** and **where** to capture that information from. | AI gives a summary of what to capture and why. It also updates the code and adds log lines in the code | Human evaluates the LLMs output and determines if the suggestions are correct and if not, guides the LLM towards a better answer | - Familiarity with the system and its expected behavior. | - Tech stack<br>- System architecture<br>- Design patterns | - Observant<br>- Transparent | - LLM Integrated code editor<br>- LLM integrated debugging tools |
| 11C | **Implement tracing (e.g., adding tracing calls to use Open Telemetry library) for critical flows** | *Example: I want to save guest info into the database in a hotel reservation system. Along the process, I don't know what might go wrong. I ask an LLM to generate a tracing component (for example, a piece of code from an API) that can check if hotel rooms are available while it is trying to save credit card information. Tracing is related to structured data formats.* | Gives AI access to the code and prompts it to update code to enable more detailed tracing for requests related to the CUJ. | AI updates code to enable the more detailed tracing and details how to access and interpret the results. | Human approves code updates, and takes code live either locally or in production. Tracing data is now available to both human and AI contributors to aid in future reliability optimization. | - Familiarity with potential tracing options and how to interpret and act on the tracing data produced. | - Tech stack<br>- System architecture<br>- Design patterns | - Observant<br>- Time Consistent | - LLM Integrated code editor<br>- LLM Integrated tracing tool |
| 11D | **Plan for contingencies** | *Example: I need to plan for disaster, capacity, redundancy and rollbacks. I prompt AI to ask me what I need to consider my scenario. Say I am going to add a really big customer, AI helps look at logs and figure out what else I need to integrate the customer to my system. Ask LLMs what could possibly go wrong for a new client? What are the points of this plan and give me an outline to solve it.*<br><br>*Example: writing an emergency response playbook.* | Human prompts AI trained on internal resources for a risk analysis document or premortem related to a planned launch. | AI enumerates and ranks scenarios based on current code and documents as well as historical precedents and produces the requested document based on any identified plausible risks. | Human iterates on the document with further AI input, adjusting both launch plan and risk document. | - Plan project<br>- Prioritize<br>- Determine trade-offs (prioritizing what to remediate after identifying points of failure)<br>- integrate human + AI knowledge | - Tech stack<br>- System architecture<br>- Tech debt | - Observant<br>- Perfectionist<br>- Systematic | - LLMs<br>- Collaboration tools<br>- Project mgmt tools<br>- Test environments |

# 12. Produce up-to-date documentation

**Note:** For each of the tasks in this section, the cross-cutting SKAs from the previous section apply, in addition to the SKAs specific to each task.

## 12. Produce up-to-date documentation

**Note:** For each of the tasks in this section, the cross-cutting SKAs from the previous section apply, in addition to the SKAs specific to each task.

| | Tasks | Examples | What the human does | What the AI does | What the human does | Skills | Knowledge | Attributes | Tools |
|---|---|---|---|---|---|---|---|---|---|
| 12A | **Assess whether changes warrant any documentation and identify gaps** | *Example: When fixing a bug, with only API public documentation. Ask AI if any changes are referenced in the documentation to make the change.*<br><br>*Example: Ask an LLM to recognize incongruencies from changes made to the code* | The human provides LLM with the outdated docs and updated code to ask the model to update the docs with changes from the code. | The AI identifies various outdated docs and code samples. Then AI either suggests the changes or returns updated docs. | The human assesses the identified locations to be changed for relevance and/or correctness. The human iterates with AI until he or she are satisfied with the output. | - Conduct research / inquiry<br>- Technical writing<br>- Reading documentation | - Existing documentation and scope/detail therein | - Observant | - LLM (e.g., ChatGPT, Gemini), Github |
| 12B | **Document throughout the process** | *Example: **Ask Co-Pilot** to automatically add comments to your code as you go. You tell it to holistically create new documentation for something you've written if it does not already exist.* | The human, as they are writing code, notices blocks or lines of code that may not be immediately obvious to the reader, or wants to document a logical step or caveat being mitigated and types the single character to begin a comment. | The AI automatically suggests a comment that matches the following the logic that you (or it!) has written | Checks that the comment matches what they wanted to write and accepts or rejects the auto-complete. Occasionally, slight editing is needed and AI may be leveraged to auto-complete sentences. | - Technical writing | - Tech stack | - Efficient<br>- Empathetic<br>- Organized | - LLM (e.g., ChatGPT, Gemini), Github |
| 12C | **Make changes in all documentation needed** | *Example: Ask AI to **identify** all locations to make changes in documentation (eg. public, internal, release notes, quick setup, known issues)*<br><br>*Example: Take all the changes (e.g. pull requests, commit history, etc) you've made in GItHub and export it into a file to use an LLM to generate documentation*<br><br>*Example: If you removed a variable that no longer exists in your code, Ask AI to remove references to that variable in documents* | The human gives AI access to the area(s) of documentation and provide changes made in relevant work (eg. Pull Request). Prompt AI to find all relevant documentation in need of updating due to the changes contained therein. | The AI understands what has changed and where that is mentioned within all sources of documentation and suggest changes to documentation | Review changes to documentation and implement changes. | - Technical writing | - Existing documentation and scope/detail therein<br>- Feature being worked on / changed<br>- Intended audience | - Observant<br>- Organized | - LLM (e.g., ChatGPT, Gemini), Github<br>- Special-purpose AI tool / app |
| 12D | **Generate example code for tutorials** | *Example: Ask an LLM to generate example code written with the latest example of a programming language library.* | The human gives AI access to the library and any documentation. The human prompts the AI to generate small samples of code that utillize the library, ensuring to cover all edge cases catered to inside the code and using the minimum required params as well as each optional parameter or combination of optional parameters as covered by the logic of the authored library. | Generates example code snippets for various use cases. | Takes the example code snippets and adds them supplemental to or inline with the documentation, as relevant. | - Technical writing | - Domain knowledge of user-facing application and tools<br>- Tech stack | - Observant<br>- Organized | - LLM (e.g., ChatGPT, Gemini), Github |

# 13. Test the AI model

**Note:** For each of the tasks in this section, the cross-cutting SKAs from the previous section apply, in addition to the SKAs specific to each task.

# 13. Produce up-to-date documentation

**Note:** For each of the tasks in this section, the cross-cutting SKAs from the previous section apply, in addition to the SKAs specific to each task.

| | Tasks | Examples | What the human does | What the AI does | What the human does | Skills | Knowledge | Attributes | Tools |
|---|---|---|---|---|---|---|---|---|---|
| 13A | **Obtain the expected/labeled inputs and outputs (or datasets)*** | *Example: I am building a self-driving car model and need the car to learn how to react to inputs. For example, how to behave (output) if there's a delivery truck standing in front of it (input)? I need to test my model against all of these various inputs. AI synthesizes and captures an envelope of behavior for this model along all of these input and output sets.* | Provide to the AI the desired/acceptable input/output pairs that is critical to the functioning of the system. Provide with a few functional examples | AI classifies or outputs a **detailed set of outputs and inputs**, **including negative inputs** that warrant the system to reject those inputs. | Human then validates the response by reviewing the input and output sets, and discarding the sets that do not make sense or out of the envelope of behavior of the system | - Use special-purpose AI model or app.<br>- Test the special-purpose AI model or app (for functionality)<br>- Evaluate predictions / classifications made by special-purpose AI model or app. (for bias, for example) | - Design patterns | - Confident<br>- Perfectionist<br>- Proud of your work | - Special-purpose AI Model or App |
| 13B | **Validate the expected/labeled inputs and outputs (or datasets) with real world knowledge*** | *Example: I need to validate my system's input sets to weed out what is inappropriate or not applicable, so I use AI to prune the data.*<br><br>*Example: If you are rolling out a self-driving car product in California you don't want the input dataset to be from India where the traffic patterns are very different.* | Provide to the AI the desired/acceptable input/output pairs and the metadata about the function. This might include environment information, metadata, previous usage and behavior characteristics | AI based on the real world knowledge of the input artifacts, validates the input-output or behavior sets to **find anomalies** in behavior patterns, especially when gauged against the exact usage scenarios. AI **flags** out the behavior patterns of the model, which are very different to the real working condition of the system | Human then validates the response, and reviews the suggestions for correctness such that there are **not false positives or true negatives in the suggested validations**. | - Use special-purpose AI model or app.<br>- Test the special-purpose AI model or app (for functionality)<br>- Evaluate predictions / classifications made by special-purpose AI model or app. (for bias, for example) | - Design patterns<br>- Efficiency of the model / product | - Perfectionist | - Special-purpose AI Model or App |
| 13C | **Execute the model or application with input set to generate outcomes*** | *Example: An AI system designed to drive a car decides how to behave given a series of inputs such as a person taking a long time to cross the road.*<br><br>*Example: Browser agent mimics work performed by humans and can book a flight, schedule a meeting, etc.* | Provides to the AI the **context** of the **system** along with **input** & **output** sets. The context would be learned sequentially and the LLM model would be supplied with inputs, performance and functional envelope of the system | AI **runs the model with** the **input** that are supplied and **generated to compute purported outcomes** for the system. AI then supplies the results to the user in the form of outcomes. | Human, while validating the outcomes, reviews the outcomes to find any anomalies that need retraining the model, rethinking the system or rethinking the inputs over iterations. After a few iterations, the outcome produced will be exactly as desired and defined | - Use special-purpose AI model or app.<br>- Test the special-purpose AI model or app (for functionality)<br>- Evaluate predictions / classifications made by special-purpose AI model or app. (for bias, for example) | - Application / Product / Model<br>- Software product design<br>- System architecture<br>- Various data formats<br>- Design patterns | - Observant<br>- Proud of your work | - Special-purpose AI Model or App<br>- Data Analysis tools / dashboards<br>- Interpreter environments<br>- Browser / web agents |
| 13D | **Compare model or application behavior against intended behavior*** | *Example: Ask AI to capture and verify match score over various sizes and attributes. Consider how far from the intended response was.* | The human provides the **test performed** along with the application response and **expected** behavior as context. The human also provides as context the **rubric** with which to evaluate the outcomes of the model/application. With this context, they prompt the **AI to grade each outcome** and **flag discrepancies** between expected and obtained outputs | The AI **generates the scores for each test performed** and provides a **summary** description regarding the performance of the test | The human can then verify the test results focusing on the ones flagged by the AI | - Use special-purpose AI model or app.<br>- Test the special-purpose AI model or app (for functionality)<br>- Evaluate predictions / classifications made by special-purpose AI model or app. (for bias, for example) | - Application / Product / Model | - Observant<br>- Perfectionist | - Special-purpose AI Model or App<br>- Data analysis tools / dashboards |

## 13. Produce up-to-date documentation

**Note:** For each of the tasks in this section, the cross-cutting SKAs from the previous section apply, in addition to the SKAs specific to each task.

| | Tasks | Examples | What the human does | What the AI does | What the human does | Skills | Knowledge | Attributes | Tools |
|---|---|---|---|---|---|---|---|---|---|
| 13E | **Publish accuracy scores aggregated across experiments** | *Example: Ask AI to produce a code to analyze the data to yield the intended aggregate result. - E.g. 90% of the time, Waymo performed with more than 95% accuracy.*<br><br>*Example: Ask LLM to explain the things that are going wrong for there to be a discrepancy between expected and obtained outcome* | The human provides all of the test results and scores generated as context and prompts the AI to generate summary statistics. | The AI processes the information provide to generate a **report** describing the overall performance of the model/application | The human reviews the report and can iterate further with the AI **asking for explanations** of specific performance metrics | - Use special-purpose AI model or app.<br>- Test the special-purpose AI model or app (for functionality)<br>- Evaluate predictions / classifications made by special-purpose AI model or app. (for bias, for example) | - Tech stack<br>- Software product design<br>- System architecture | - Observant<br>- Perfectionist | - LLM or LMM<br>- Data analysis tools / dashboards |

# Acknowledgements

# Acknowledgements (All names organized in alphabetical order by last name.)

| Project Team | Panelists | Google Advisors |
|---|---|---|
| Erin Barrar | Nikhil Gupta | Simeon Anfinrud |
| Joseph Ippito | Ashwin Hegde | Michael Dedrick |
| Matthew Kam | Kartheek Kolla | Andrew Gleeson |
| Joshua Kenitzer | Cha Li | Noah Jadallah |
| Irene A. Lee | Casey Pollock | Josh Kahn |
| Joyce Malyn-Smith | Russell Preston | Thomas Keck |
| Cody Miller | Suparna Srinivasan | Chris Liu (Xi) |
| Beatriz Perret | Vikram Tiwari | Paul Macovei |
| Abey Tidwell | | Adrian Matteis |
| Miaoxin Wang | | Aleks Rozman |
| | | Joe Simunic |
| | | Kevin Wiesmueller |
| | | Celal Ziftci |

**For details of the adapted DACUM process or our commentary on the AI-enhanced software developer described in this occupational profile:**

Google

**Google Contact**
Matthew Kam
mattkam@google.com

EDC.ORG

**EDC Contact**
Joyce Malyn-Smith
jmsmith@edc.org